\documentclass[10pt,twocolumn,letterpaper]{article}

\usepackage{booktabs}
\usepackage{multirow}

\usepackage{graphicx}
\usepackage{amsmath}
\usepackage{amssymb}
\usepackage{cuted}
\usepackage{makecell}
\usepackage{bbding}
\usepackage{mathbbol}
\usepackage{amsmath}

\usepackage{tabularx} % for 'tabularx' env. and 'X' col. type
\usepackage{ragged2e} % for \RaggedRight macro
\usepackage{booktabs}
\usepackage[accsupp]{axessibility}
\usepackage[pagenumbers]{arxiv} % To force page numbers, e.g. for an arXiv version

\usepackage[dvipsnames]{xcolor}

\definecolor{arxivblue}{rgb}{0.21,0.49,0.74}
\usepackage[marginal]{footmisc}
\usepackage[pagebackref,breaklinks,colorlinks,citecolor=arxivblue]{hyperref}
\newcommand{\myparagraph}[1]{\vspace{0.1em}\noindent\textbf{#1}}

\title{BOTH2Hands: Inferring 3D Hands from Both Text Prompts and Body Dynamics}

\author{Wenqian Zhang,\ Molin Huang,\ Yuxuan Zhou,\ Juze Zhang,\ Jingyi Yu,\ Jingya Wang,\ Lan Xu\\
ShanghaiTech University\\
{\tt\small \{zhangwq2022,huangml,zhouyx2,zhangjz,yujingyi,wangjingya,xulan1\}@shanghaitech.edu.cn}
}
\begin{document}
\maketitle
% \begin{figure*}
\begin{strip}\centering
\includegraphics[width=1\linewidth]{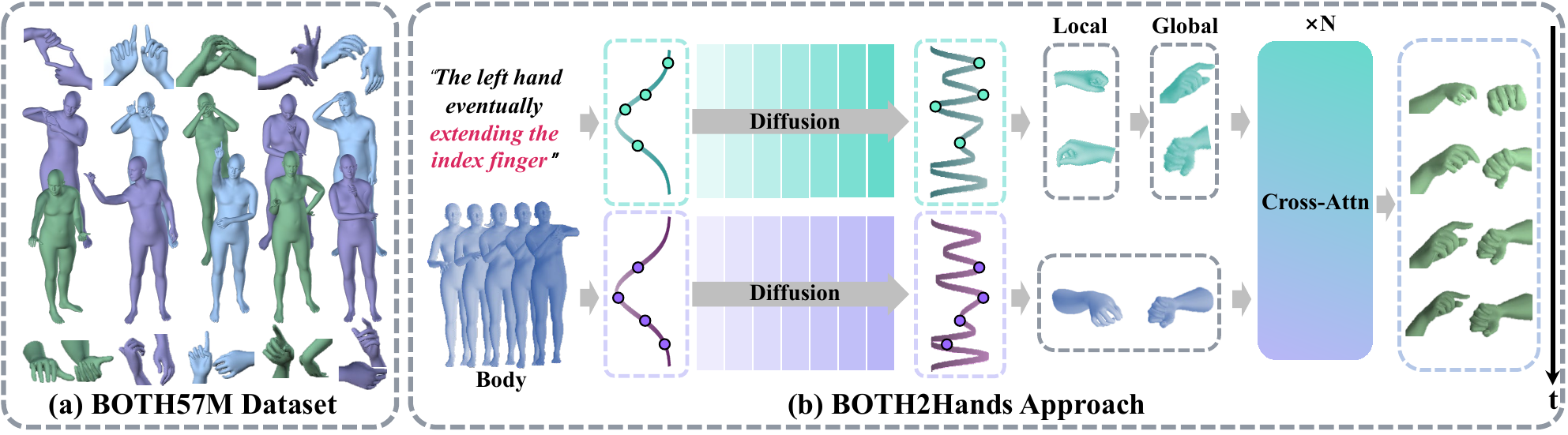}
\captionof{figure}{(a) Our BOTH57M dataset contains rich gestures and body movements. (b) BOTH2Hands is the only model that can handle text prompts and body dynamics as input, generating realistic hand motions at present.}
\label{fig1:teaser}
\end{strip}
% \end{figure*}

\begin{abstract}
The recently emerging text-to-motion advances have inspired numerous attempts for convenient and interactive human motion generation. Yet, existing methods are largely limited to generating body motions only without considering the rich two-hand motions, let alone handling various conditions like body dynamics or texts. To break the data bottleneck, we propose BOTH57M, a novel multi-modal dataset for two-hand motion generation. Our dataset includes accurate motion tracking for the human body and hands and provides pair-wised finger-level hand annotations and body descriptions. We further provide a strong baseline method, BOTH2Hands, for the novel task: generating vivid two-hand motions from both implicit body dynamics and explicit text prompts. We first warm up two parallel body-to-hand and text-to-hand diffusion models and then utilize the cross-attention transformer for motion blending. Extensive experiments and cross-validations demonstrate the effectiveness of our approach and dataset for generating convincing two-hand motions from the hybrid body-and-textual conditions. Our dataset and code will be released to the community for future research, which can be found at \href{https://github.com/Godheritage/BOTH2Hands}{github}.
\end{abstract}

\section{Introduction}
\label{sec:intro}

The recent years have witnessed the tremendous progress of human motion generation, especially for the recently emerging text-to-motion setting~\cite{tevet2023human,tevet2022motionclip,zhang2023t2m,chen2023executing}. It enables novices to conveniently generate desired motions in a natural interactive manner. Yet, realistic human motions require the generation of companion motions of hands. Actually, we humans tend to incorporate a wide variety of hand motions with body movements in our daily communications.

The recent text-to-motion advances~\cite{tevet2023human,liang2023intergen,dabral2023mofusion,zhang2023t2m} mostly focus on generating body motions only. In contrast, the convenient generation of two-hand motions from text prompts has significantly fallen behind, mainly due to severe data scarcity. The wider adopted text-motion datasets~\cite{mahmood2019amass,guo2022generating} embrace limited two-hand motions and corresponding textual annotations.
Only recently, the concurrent work Motion-X~\cite{lin2023motion} provides a large-scale dataset with expressive human motions and paired text prompts. 
Yet, it still lacks detailed annotations for the hand motions, making the fine-grained generation challenging, let alone enabling explicit finger-level controls. On the other hand, various methods~\cite{ng2021body2hands,qi2023diverse,qi2023emotiongesture} synthesize two-hand motions with body motion as extra conditions. Such a body-to-hand setting implicitly reasons the inherent correlations of human motions between the body and hands, and hence effectively handles specific scenarios like speeches or daily conversations~\cite{zhu2023taming,yin2023emog}. However, only body-level reasoning falls short of providing explicit and direct controls of hand motions, especially in a human-interpretable manner like text prompts.

To tackle the above challenges, in this paper, we present \textit{BOTH2Hands} -- a novel scheme to generate two-hand motions under a novel and hybrid setting: from both text prompts and body dynamics, as illustrated in Fig.~\ref{fig1:teaser}. By organically combining the explicit and implicit conditions, our approach enables vivid and fine-grained hand motion generation. Nevertheless, generating hand motions in such a novel setting is challenging. First, it requires fusing and balancing the conditions from two very different modalities~\cite{karunratanakul2023guided}, which may point to diverse generation results. Second, the fundamental data scarcity for two-hand motion generation remains, while such a novel multi-modal further constitutes barriers to data annotations.

Specifically, we first introduced a large-scale multi-modal dataset, named \textit{BOTH57M}, for two-hand motion modeling.  Our dataset includes accurate hands and body motions with paired finger-level hand annotations and body descriptions, under diverse activities, covering 57.4 million frames of 8.31 hours with 23,477 textual annotations. To handle the occlusion, we adopt a camera dome with 32 RGB input views and utilize the off-the-shelf motion capture approach~\cite{he2021challencap} to faithfully recover the skeletal motions of both the hands and body. We then provide two types of textual annotations for the captured motions: one describes the full body motions in general, while another focuses on fine-grained hand motions with finger-level and highly precise annotations. Note that our BOTH57M dataset is the first of its kind to open up future research for two-hand motion generation under hybrid conditions of both body dynamics and text prompts. Our accurate motions and expressive annotations also bring substantial potential for future direction in multi-modal control or human behavior analysis.

Based on our novel dataset, we further propose BOTH2Hands, a strong baseline approach to generate vivid two-hand motions from diverse conditions like body motions and text prompts. We tailor the recent diffusion models~\cite{ho2020denoising} into a two-stage mechanism for this novel task. Our core idea is to optimize the potential of the diffusion model using each modality separately and subsequently utilize a cross-attention transformer to blend them into a two-hand motion generation with multi-conditioning. Specifically, we warm up two parallel body-to-hand and text-to-hand diffusion models in the first stage. Then, we leverage a cross-attention transformer for motion blending, where two conditioned results are alternately inserted into the attention layers to generate convincing and vivid two-hand motions. Finally, we present a thorough evaluation of our approach against various state-of-the-art motion generation methods using our dataset. We also perform cross-validation on both our dataset and the concurrent Motion-X~\cite{lin2023motion} dataset, demonstrating the enhancement of our dataset for the two-hand generation task.
To summarize, our main contributions include:
\begin{itemize}

\setlength\itemsep{0em}

    \item  We propose a novel scheme to generate fine-grained two-hand motions under a novel setting: from both implicit body dynamics and explicit text prompts.
    
    \item  We contribute a large-scale multi-modal dataset for two-hand generation, with accurate body and hand motions as well as rich finger-level textural annotations.
    
    \item We combine parallel diffusion structures with a subsequent cross-attention transformer, to effectively generate hand motions from various conditions.

    \item To tackle the data scarcity, we will release our dataset, codes, and pre-trained models for future exploration.

\end{itemize}

\begin{figure*}[t!]
  \centering
  \includegraphics[width=1\linewidth]{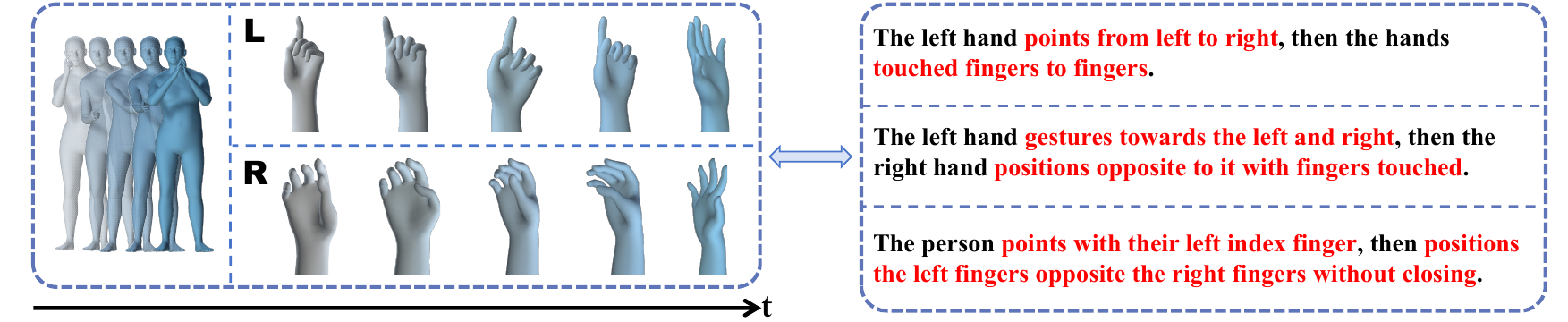}
  \vspace{-8mm}
  \caption{BOTH57M dataset focuses on body-hand motions within the daily scene, incorporating a vast and versatile collection of daily gestures. Each segment of hands within the dataset has been manually annotated three different times. For more details, please refer to the supplementary materials.}
  \label{fig:dataset}
  \vspace{-8pt}
\end{figure*}

\begin{table*}[t]
\centering
\begin{tabular}{l|l|l|l|l|l|l|lll} 
\hline
Dataset & \#Frame & \#Cam-view & Vocab & RGB & Annotations & Hand Joints std $\uparrow$  & \multicolumn{2}{c}{\#Detailed annotation}  \\
       &       &      &       &     &  & & Body & Hand  \\ 
\hline
GRAB    & 1.6M & 54   & $\times$ &  $\times$  & $\times$ & 0.675   & $-$   & $-$     \\
EgoBody  & 220K        & 6        & $\times$            &  $\checkmark$   & $\times$       &  0.316         & $-$     & $-$                       \\
BEAT    & 32M &  16       & $-$           &\checkmark     &  $-$       &  0.076         & $-$     & $-$                       \\
MotionX &  13.7M       &  $-$       &  2898    & \checkmark    &  \checkmark       &  0.186         & 1     & 1                        \\ 
\hline
BOTH57M(Ours)    &  57.4M       &  32       &  4140      & \checkmark    &  \checkmark       &   0.422        &3      &3                        \\
\hline
\end{tabular}
\vspace{-3mm}
\caption{\textbf{Dataset comparisons.} We conduct a comparison of datasets that encompass body and hand motions. \textbf{Vocab.} denotes the distinct vocabulary numbers used for annotation. \textbf{Annotation} refers to text annotations. \textbf{Hand Joint Standard Deviation} reflects the standard deviation of hand joint positions, indicating the diversity of hand motions. \textbf{Detailed annotation} refers to the number of text annotations for specific skeleton parts in each motion clip.}
\label{tab:dataset}
\vspace{-15pt}
\end{table*}

\section{Related Works}
\label{sec:related}
\myparagraph{Motion Generation.} 
Currently, numerous works focus on motion generation under various conditions such as text and label~\citep{tevet2023human,zhang2023t2m,zhang2022motiondiffuse,plappert2018learning,plappert2016kit,guo2022generating,delmas2022posescript,ahuja2019language2pose,zhao2023modiff,petrovich2021action,athanasiou2022teach},
speech and music ~\citep{yi2023generating,liu2022audio,zhu2023taming,ye2022audio,habibie2021learning,zhao2023taming,yao2023dance} and
objects~\citep{taheri2020grab,brahmbhatt2020contactpose,corona2020ganhand,hasson2019learning}. Other interesting works use brand-new algorithms~\cite{he24nrdf} or focus on new scenes~\cite{lin2023handdiffuse}.
Among these, text-to-motion generation is a challenging task due to the difficulty of aligning natural language with time and space~\cite{petrovich2022temos,kim2023flame}. 
MotionClip~\cite{tevet2022motionclip} aligns text with other modalities, enhancing model mapping text to motion. As diffusion model~\citep{song2020denoising,ho2020denoising,song2021scorebased,dhariwal2021diffusion} was introduced in various tasks~\citep{rombach2022high, sarandi2023learning,dabral2023mofusion,rombach2022high,saharia2022photorealistic}, it also performs well in motion generation. For instance, Human Motion Diffusion Model~\cite{tevet2023human} introduced text conditions and showed good results. 
Other works like T2M-GPT~\cite{zhang2023t2m} applied the transformer architecture in this task and proved its effectiveness. MLD (motion-latent-diffusion)~\cite{chen2023executing} has attempted to generate motions in the latent space. Some other works like InterGen~\cite{liang2023intergen} focus on human interaction scenes achieving good results. Full-body motion generation holds considerable significance in some specific domains like human object interaction (HOI)~\citep{taheri2020grab,ghosh2023imos,wu2022saga,li2024task,tendulkar2023flex} and speech~\citep{yi2023generating,ao2023gesturediffuclip,liu2022learning,yang2023qpgesture}, due to its extensive applicability.
%{Unifying}

Hands are equally important as bodies in motion~\cite{muller2023generative}, presenting unique challenges due to their high density in small spatial occupancy. Previous hand generation works concentrate on physics-based issues~\cite{pollard2005physically,liu2008synthesis,zhao2013robust} such as surface contact~\cite{zhang2021manipnet} and reconstructing hands~\cite{romero2017embodied,li2022nimble}. Other data-driven~\cite{jorg2012data,majkowska2006automatic,mousas2015finger,stone2004speaking} works often focus on specific scenarios~\cite{zhang2023neuraldome}, modeling the individual as a whole rather than considering parts separately~\cite{fieraru2020three,xu2023h2onet}. Some works like Body2Hands~\cite{ng2021body2hands} take body as condition and achieve impressive results~\cite{qi2023diverse,qi2023emotiongesture,yin2023emog,zhu2023taming}. Previous studies often focus on full-body or body motions only, rather than generating hands aligned with both body motions and text controls.

\myparagraph{Motion Dataset.}
These days various motion datasets have been presented. Action-labeled datasets like BABEL~\cite{punnakkal2021babel} offer verb-object phrases as conditions, which is unnatural for human communication. Datasets such as KIT~\cite{plappert2016kit} or HumanML3D~\cite{guo2022generating} provide detailed natural annotations, while they ignore hands. Other datasets focus on hand scenarios like Hand-Object Manipulation~\cite{fan2023arctic}, 3D Interacting Hand~\cite{Li_2023_ICCV,moon2020interhand2}. Yet, such scenes mostly focus on hands, they hardly contain both hand and body data.
Full-body datasets like GRAB~\cite{taheri2020grab} contain rich hand gestures but are narrowed down to HOI scenes. BEAT~\cite{liu2022beat} uses speech text as conditions, lacking standard motion description. Currently, the largest full-motion dataset Motion-X~\cite{lin2023motion} contains descriptions aligned with motions but lacks annotations focusing on hands, and the diversity of hand movement is less rich than their body motions. More data including rich daily hand gestures with detailed annotation is needed for body hand motion synthesis.

\section{BOTH57M Dataset} \label{sec:dataset}

\myparagraph{Overview.}
We introduce the BOTH57M, a unique body-hand motion dataset comprising 1,384 motion clips and 57.4M frames, with 23,477 manually annotated motions and a rich vocabulary of 4,140 words. The dataset focuses on hands and body motion in daily various activities, referencing the book \textquotedblleft Dictionary of Gestures \textquotedblright ~\cite{caradec2018dictionary} and supplementing with custom-designed movements. To the best of our knowledge, this is the only dataset that provides hybrid and detailed annotations of both body and hands at present, providing huge potential for future research. Tab.~\ref{tab:dataset} shows a detailed comparison of various body hand datasets with ours. The rich vocabulary and hand diversity underscores our advantage in tackling the text/body-to-hand task.

\myparagraph{Data Collecting.}
We utilize 32 RGB cameras to build a dense-view system for body-hand motion capturing. During data collection, participants are instructed to perform movements listed in the \textquotedblleft Dictionary of Gestures\textquotedblright \ excluding unfriendly gestures. Subsequently, manual annotations are implemented. Three annotators are required to annotate full-body motions. For hand motions, three other annotators individually annotate finger-level actions for the left and right hand, focusing on the changing process of finger movements and gaining detailed records for prominent finger gestures. Fig.~\ref{fig:dataset} offers a comprehensive exemplification of our dataset. For a comprehensive understanding of data collection and processing, as well as an in-depth explanation of our collected motions and text annotations, please refer to the supplementary material.

\begin{figure*}[t!]
  \centering
  \includegraphics[width=1.0\linewidth]{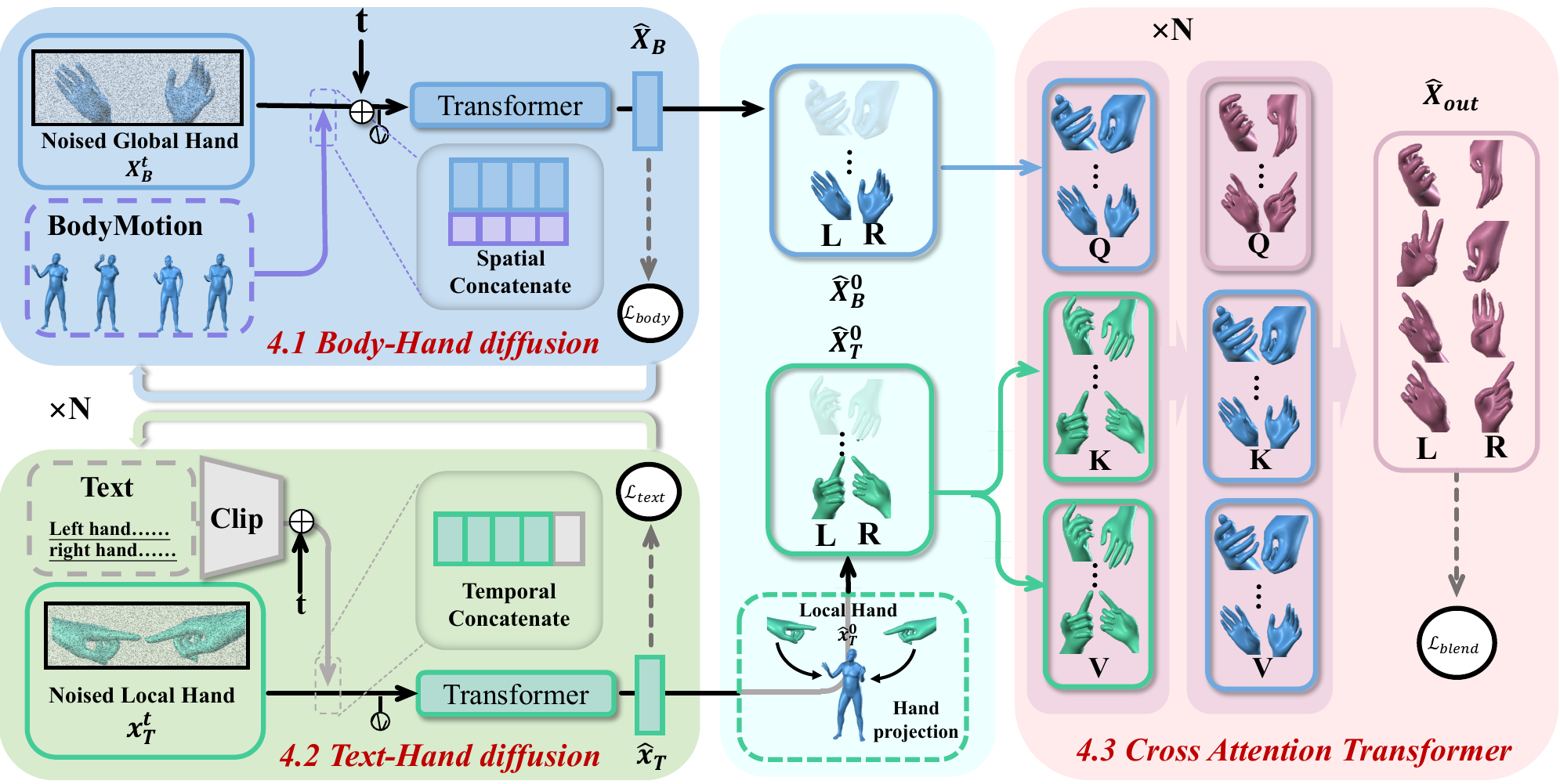}
  \vspace{-5mm}
  \caption{\textbf{Overview of BOTH2Hands pipeline.} Our pipeline initially feeds text prompts and body movements into two separate diffusion models. Subsequently, the text-conditioned outcomes are projected into the body-conditioned hand coordinate system using forward kinematics. Finally, we utilize the two sequences of hand motions as inputs into a cross-attention transformer for motion blending.}
  \label{fig:pipeline}
  \vspace{-5mm}
\end{figure*}

\section{BOTH2Hands Algorithm}
\label{sec:method}
Based on our novel dataset, our objective is to generate hand motions that align with both textual prompts and body movements. To accomplish this, we propose a novel pipeline called BOTH2Hands to deal with rich conditions to generate lively two-hand motions, as shown in Fig.~\ref{fig:pipeline}. Our framework consists of a two-stage mechanism: a pair of diffusion-based hand motion denoisers and a cross-attention structured transformer. In the first stage, we feed the body and text controls into two parallel diffusion models. In the second stage, we blend the hand motions generated by two-modality controls. Specifically, we follow EgoEgo~\cite{li2023ego} and adopt forward kinematics (FK) to get joint positions and rotations 6D in body motion space. For body-conditioned hand diffusion, our goal is to generate hand motions that are coordinated with the body motions (Sec.~\ref{sec: Body-hand motion diffusion}). As for text-conditioned hand diffusion, we first use inverse kinematics (IK) to get local hands to make the denoise process more focused on gesture. However, these generated hands are not in the same coordinate system as the body-conditioned hands. To address this issue, we then employ FK to project the local hands back into the body motion space, thereby eliminating gesture rotation errors while blending (Sec.~\ref{sec: Text-hand motion diffusion}). After that, we perform cross-attention motion blending between the text-conditioned hands and the body-conditioned hands (Sec.~\ref{sec: Cross-attention based gesture blending}). This process ensures the generated hand motions effectively combine the dynamics of body motion with the explicit textual conditions.

\subsection{Body-hand motion diffusion} \label{sec: Body-hand motion diffusion}

\myparagraph{Motion Diffusion Model.}
In our approach, we adopt the formulation suggested in the denoising diffusion probabilistic model (DDPM)~\cite{ho2020denoising}, which effectively handles the hand synthesis task. 
The diffusion model processes the input data ($\textbf{x}_{0}$) for $t$ iterations and obtains the noised data at level $t$. In each iteration, sampled Gaussian noise is added to the data from the previous level.
This iterative process is commonly referred to as the forward process and can be represented as a Markov chain with $t$ steps. The transition probability is shown in Eq.~\ref{eq:Markov}:
\begin{equation} \label{eq:Markov}
\begin{aligned}
q(x_t|x_{t-1})=\mathcal{N}(\sqrt{\alpha_t}x_{t-1},(1-\alpha_t)\mathbf{I}),
\end{aligned}
\end{equation}
where $\beta_{t}$ is a variance schedule parameter and $\alpha_{t} = \prod_{i=1}^t{(1- \beta_{i})}$. The reverse diffusion process can be modeled as $p_{\theta}(x_{t-1}|x_t,c_{0:N})$, where $\theta$ represents the learned parameters and $c_{0:N}$ represents a set of given conditions ($0$ indicates no condition). Notable, we can always train a diffusion denoiser with any condition to learn a Gaussian posterior distribution $q(x_{t-1}|x_t,x_0)$. 
The denoising sampling process can be formulated as:
\begin{equation} \label{eq:BackMarkov}
\begin{aligned}
p_{\theta}({x}_{t-1}|{x}_{t}, c_{0:N}) = \mathcal{N}({x}_{t-1}; {\mu}_{\theta}({x}_t, t, c_{0:N}), \sigma_{n}^{2}\mathbf{I}).
\end{aligned}
\end{equation}
The term $\mu_{\theta}({x}_t, t, c_{0:N})$ is the mean to learn, which can be impacted by the conditions $c_{0:N}$. As mentioned in previous works~\citep{dhariwal2021diffusion,karunratanakul2023guided}, to be more exact, the updating rule of the mean is:
\begin{equation} \label{eq:MultiCondMean}
\begin{aligned}
\mu^{c_{0:N}}_{t} = \mu^{c_{0}}_t +\sum_{i=1}^{N}s_{i}(\nabla \log p(c_{i}|x_{t})),
\end{aligned}
\end{equation}
where $\mu^{c_{0}}_t$ is the mean without condition and the gradients of joint condition are noted as $\sum_{i=1}^{N}(\nabla \log p(c_{i}|x_{t}))$. The weights $s_i$ controls the strength of conditioning. However, since different conditions pertain to distinct modalities, it is hard to manually configure the strength parameter $s_{i}$. Therefore, the preference leans towards the utilization of separate diffusion models to avoid imbalanced control by different conditions. This approach prevents being affected by other conditions when learning $\mu_t$ since the gradient of a single condition is far easier to learn than joint conditions:
\begin{equation} \label{eq:SingleCondMean}
\begin{aligned}
\mu^{c}_{t} = \mu_t +\nabla \log p(c|x_{t}),
\end{aligned}
\end{equation}
where $c$ indicates one single condition. 

\myparagraph{Body to Hand Diffusion.}
As the same body motion may lead to different hand gestures, we need the diffusion probabilistic model to sample the most possible gesture instead of directly matching one gesture on the body.
We utilize the widely adopted model SMPLH (SMPL+MANO)~\cite{loper2023smpl,romero2017embodied} as our skeleton, with a total of $52$ joints, where the initial $22$ joints are body joints and the remaining $30$ joints are hand joints. We parameterize the representation of motions as positions and rotations 6D of joint. To fetch parameters, we use FK to calculate the absolute rotations and joint positions (we define absolute parameters as real positions and rotations of joints without the intervention of parent joints),
denoted as global motions, where the former $22$ joint positions and rotations 6D correspond to global body ($\textbf{C}_{B}\in \mathbb{R}^{T \times 22 \times 9}$) and the latter $30$ joints correspond to global hands ($\textbf{X}_{B}\in \mathbb{R}^{T \times 30 \times 9}$). We perform the forward process by adding noise to the hand motions step-by-step, fetching the sequences of hand motions $\textbf{X}^{t}_{1},\textbf{X}^{t}_{2},...,\textbf{X}^{t}_{T}$ at noise level $t$. Followed by the forward process, we conduct a reverse diffusion procedure on the transformer self-attention denoiser to estimate $\textbf{X}^{0}_{B}$. Then we adopt methods in~\cite{li2023ego}
to directly concatenate the noised hand $\textbf{X}^{t}_{B}$ and cleaned body $\textbf{C}^{0}_{B}$ together as denoiser input during the body-hand diffusion process, with the loss shown in Eq.~\ref{eq:SimpleLoss}:
\begin{equation} \label{eq:SimpleLoss}
\begin{aligned}
\mathcal{L}_{body} = \mathbb{E}_{{\textbf{X}}_0, t}||\hat{{\textbf{X}}}_{\theta}({\textbf{X}}^{t}_{B}, t, \textbf{C}^{0}_{B}) - \textbf{X}^{0}_{B}||_{1}.
\end{aligned}
\end{equation}
We directly predict the cleaned motion $\textbf{X}^{0}_{B}$ and use reconstruction loss as diffusion training loss.

\subsection{Text-hand motion diffusion} \label{sec: Text-hand motion diffusion}
 For text-conditioned hand synthesis, we use IK to extract the local positions and rotations 6D from FK motion results. Then we discard body rotations, keeping hand rotations 6D as ground truth, the calculation process is defined below:
\begin{equation} \label{eq:hand_global_to_local_rotation}
\begin{aligned}
\textbf{x}^{rot}_{T} = Cat(IK(\textbf{M}^{rot}_{lhand}),IK(\textbf{M}^{rot}_{rhand})),
\end{aligned}
\end{equation}
where $\textbf{M}$ is the full body motion aligned with text condition, $IK(\cdot)$ is inverse kinematics process and $Cat(\cdot,\cdot)$ is concatenate operation. For joint positions, we first use FK to calculate the $52$ absolute joint positions. Then we can obtain hand joints in the origin of the coordinate system by subtracting the positions of their respective wrist for each hand joint:
\begin{equation} \label{eq:hand_global_to_local_position}
\begin{aligned}
\textbf{x}^{pos}_{T} = Cat(\textbf{M}^{pos}_{lhand}-\textbf{M}^{pos}_{lwrist},\textbf{M}^{pos}_{rhand}-\textbf{M}^{pos}_{rwrist}).
\end{aligned}
\end{equation}
We define these rotation and position groups as local hands parameters ($\textbf{x}_{T}\in \mathbb{R}^{T \times 30 \times 9}$). 

\myparagraph{Hand Projection.} 
Relative positions (we define relative parameters as joint positions and rotations relative to parent joints) representation may result in motion drifting due to the need for integrating velocity to obtain absolute positions~\cite{von2017sparse}. Nevertheless, relative rotation representations are advantageous for focusing on gestures and are easy to migrate~\cite{loper2023smpl}. Remember that we use motion representation, consisting of positions and rotations 6D. For body-conditioned hand synthesis, absolute positions and rotations of hand joints with body joints are directly applicable. %怪
However, predicting the absolute pose of hands without wrist positions is hard and meaningless for text-conditioned synthesis, so for the positions, we prefer absolute representation in the origin of the coordinate system to avoid integration prediction and parent skeleton influence. To focus on the gesture itself, we prefer to use relative rotations. Nonetheless, absolute positions and relative rotations cannot be used for hand blending directly, since the two conditioned hands are on different coordinate systems. In order to mitigate the influence of the spatial reference system, we projected $\textbf{x}_{T}$ in local space to $\textbf{X}_{T}$ in global space to eliminate their potential space error:
\begin{equation} \label{eq:text hand fk}
\begin{aligned}
\textbf{X}_{T} = FK(Cat(IK(\textbf{C}_{B}),\textbf{x}_{T})).
\end{aligned}
\end{equation}
$FK(\cdot)$ is forward kinematics process, $\textbf{x}_{T}$ are text-conditioned hand motions in local space, while $\textbf{X}_{T}$ are text-conditioned hand motions in global space.

\myparagraph{Text to Hand diffusion.}
On the text-conditioned diffusion process, we follow the methods proposed in MDM~\cite{tevet2023human} to add the text condition token $\textbf{c}$ to embed noise $t$ step token. The denoising structure is similar to the body-conditioned motion denoiser, with a slight difference in input dimension. We only feed the noised hands as input since text-conditioned synthesis can not contain body motion. We adopt the reconstruction loss similar to Eq.~\ref{eq:SimpleLoss} to predict $\textbf{x}^{0}_{T}$:
\begin{equation} \label{eq:TextSimpleLoss}
\begin{aligned}
\mathcal{L}_{text} = \mathbb{E}_{{\textbf{x}}_0, t}||\hat{{\textbf{x}}}_{\theta}(\textbf{x}^{t}_{T}, t, \textbf{c}) - \textbf{x}^{0}_{T}||_{1}.
\end{aligned}
\end{equation}

\subsection{Cross-attention hand blending} \label{sec: Cross-attention based gesture blending}
Inspired by the success of the sharing-weights transformer in InterGen~\cite{liang2023intergen}, we adopted a cross-attention transformer for gesture blending. The networks are fed with two conditioned hand motions, $\textbf{X}_{T}$ and $\textbf{X}_{B}$. We sequentially apply hand motions as attention inputs to the transformer, and compute the weighted reconstruction loss between the final output and two types of gestures. Specifically, the $\textbf{X}_{T}$ and $\textbf{X}_{B}$ are firstly embedded into a common latent space and positionally encoded into the latent states $\textbf{h}_{text}$ and $\textbf{h}_{body}$. Then, it is processed by $N$ attention-based blocks to obtain the blending hidden states $\textbf{h}^{N}_{out}$. Each block consists of multi-head cross-attention layers ($Attn$) followed by one feed-forward network ($FF$). For the first time the hands passing through the cross-attention block, the input hidden layer $\textbf{h}_{body}$ is embedded into the query matrix ($\textbf{Q}$); the attention hidden layer $\textbf{h}_{text}$ is embedded into a key matrix ($\textbf{K}$) and value matrix ($\textbf{V}$); finally we embed results into a vector $\textbf{h}^{(1)}_{out}$. The hand-blending process is detailed below:
\begin{equation} \label{eq:FirstAtten}
\begin{aligned}
&\textbf{h}^{(1)}_{out} = Attn(\textbf {Q}^{(0)},\textbf {K}^{(0)},\textbf {V}^{(0)})
=softmax(\frac{\textbf Q \textbf K^T}{\sqrt{D}}) \textbf V,\\
& \textbf{Q}^{(0)} = \textbf{h}_{body}W^{Q}, \textbf{K}^{(0)} = \textbf{h}_{text}W^{K}, \textbf{V}^{(0)} = \textbf{h}_{text}W^{V},
\end{aligned}
\end{equation}
where $D$ is the number of channels in the attention layer; $W$ are trainable weights, and $\textbf{Q}^{(i)}$, $\textbf{K}^{(i)}$, $\textbf{V}^{(i)}$ are transformer matrices under i-th layer. Passing through the attention layer once, we get the output $\textbf{h}^{(1)}_{out}$. Then we switch the $\textbf{K}$, $\textbf{V}$ input to $\textbf{h}_{body}$, which means we use body-conditioned hands as attention input to emphasize the body movements. The changed attention process is:
\begin{equation} \label{eq:SecondAtten}
\begin{aligned}
\textbf{Q}^{(1)} = \textbf{h}^{(1)}_{out}W^{Q}, \textbf{K}^{(1)} = \textbf{h}_{body}W^{K}, \textbf{V}^{(1)} = \textbf{h}_{body}W^{V}.
\end{aligned}
\end{equation}
After this, we swap $\textbf{K}$, $\textbf{V}$ input again and repeat this process until getting the final output in latent space:
\begin{equation} \label{eq:TransformerOut}
\begin{aligned}
\textbf{h}^{(N)}_{out} = FF(Attn(\textbf{Q}^{(N-1)},\textbf{K}^{(N-1)},\textbf{V}^{(N-1)})).
\end{aligned}
\end{equation}
We use blending loss to supervise the learning of weights $W$.
\begin{equation} \label{eq:Styleloss}
\begin{aligned}
\mathcal{L}_{blend} =  \mathbb{E}_{{X}_{GT}, {X}_{B},{X}_{T}}||X_{GT}-(w_{B}X_{B}+w_{T}X_{T})||_{1},
\end{aligned}
\end{equation}
where $w_{B}$ and $w_{T}$ are hyperparameters controlling weights of different hand motion parts. We set $w_{B}$ and $w_{T}$ to positive numbers and $w_{B} + w_{T} = 1$. $X_{GT}$ is GT hand motion.

\section{Experiment}
\label{sec:experiment}
We design various experiments to evaluate the validity of our method and dataset. For method evaluation, we compare our approach and baseline with existing human motion synthesis methods (Sec.~\ref{sec: Methods Evaluation}). To assess the richness and effectiveness of the BOTH57M, we train our method on training sets of BOTH57M and Motion-X separately. And subsequently evaluated trained models through the test sets (Sec.~\ref{sec: Dataset Evaluation}). Additionally, an ablation study is performed to verify the importance of hand projection and blending loss (Sec.~\ref{sec: Ablation Study}).
\begin{table*}[h]
\centering
\caption{\textbf{Quantitative evaluation} of our design with baselines and others. The \textcolor{red}{red} one and \textcolor{blue}{blue} one indicate the best result and the second best result. We use a 95\% confidence interval, approximated by the mean value plus or minus twice the standard deviation.}
\begin{tabular}{l|ccc|c|c|c|c} 

\toprule
 Methods & \multicolumn{3}{c}{R Precision$\uparrow$} & FID$\downarrow$ & MM-Dist$\downarrow$ & Diversity$\rightarrow$ & MModality$\uparrow$  \\
 \cline{2-4}
         & Top1 & Top2 & Top3&     &  &    
 &     \\ 
\hline
Real    &   $0.034^{\pm.020}$      &    $0.067^{\pm.026}$  &  $0.109^{\pm.030}$   & $0.181^{\pm.012}$ &  $1.391^{\pm.006}$     & $3.980^{\pm.090}$ &- \\
\midrule
T2M-GPT & $\textcolor{red}{\textbf{0.042}}^{\pm.014}$&$0.073^{\pm.014}$&$0.104^{\pm.020}$ & $0.461^{\pm.016}$&$1.398^{\pm.010}$&$3.689^{\pm.094}$& $1.178^{\pm.100}$ \\
MDM     & $0.039^{\pm.020}$& $\textcolor{red}{\textbf{0.077}}^{\pm.016}$&$\textcolor{blue}{\textbf{0.114}}^{\pm.024}$& $0.257^{\pm.024}$&$1.397^{\pm.008}$& $3.887^{\pm.074}$ & $1.273^{\pm.086}$ \\
MLD     & $0.036^{\pm.012}$&$0.071^{\pm.014}$&$0.106^{\pm.020}$&$0.296^{\pm.026}$&$1.400^{\pm.0014}$& $3.826^{\pm.078}$ &$1.191^{\pm.178}$ \\ 
Ego-Ego & $0.034^{\pm.022}$ &  $0.070^{\pm.030}$ & $0.109^{\pm.032}$ &$0.287^{\pm.026}$&$1.398^{\pm.012}$& $3.810^{\pm.090}$ & $1.240^{\pm.090}$ \\ 
\midrule
BOTH2Hands (Ours)  &   $0.037^{\pm.014}$     &  $0.075^{\pm.020}$   & $\textcolor{red}{\textbf{0.115}}^{\pm.028}$   &   $\textcolor{blue}{\textbf{0.201}}^{\pm.020}$     & $\textcolor{blue}{\textbf{1.392}}^{\pm.008}$     &$3.969^{\pm.082}$    &$\textcolor{red}{\textbf{1.312}}^{\pm.034}$\\
BOTH2Hands-Text  &   $0.035^{\pm.020}$     &  $0.067^{\pm.026}$   & $0.109^{\pm.030}$   &   $\textcolor{red}{\textbf{0.198}}^{\pm.012}$     & $\textcolor{red}{\textbf{1.391}}^{\pm.006}$     &$3.980^{\pm.090}$    &$\textcolor{blue}{\textbf{1.274}}^{\pm.138}$\\
BOTH2Hands-Body &   $\textcolor{blue}{\textbf{0.039}}^{+.012}$     &  $\textcolor{blue}{\textbf{0.076}}^{\pm.024}$   & $0.112^{\pm.026}$   &   $0.203^{\pm.016}$     & $1.392^{\pm.010}$     &$3.955^{\pm.098}$    &$1.266^{\pm.120}$\\
\bottomrule
\end{tabular}
\label{tab:methodcompare}
\vspace{-10pt}
\end{table*}

\begin{figure*}[t!]
    \centering
    \includegraphics[width=1.0\textwidth]{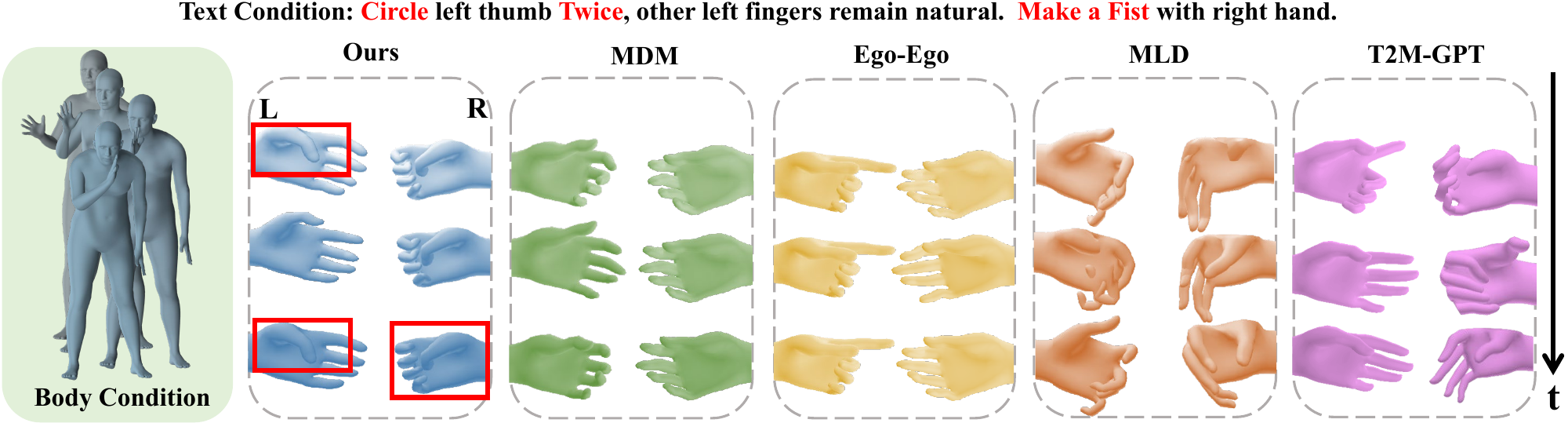}
    \caption{\textbf{Qualitative comparisons.} BOTH2Hands algorithm with other methods~\citep{tevet2023human,chen2023executing,zhang2023t2m,li2023ego} are given two conditions: text and motion. Text conditions are listed at the top, and body conditions are listed at the left side with no hands, and the temporal order is from top to bottom. The motions that follow the conditions are circled \textcolor{red}{red}.}
    \label{fig:methodeval}
    \vspace{-5mm}
\end{figure*}

\begin{figure*}[h]
    \centering
    \includegraphics[width=1\textwidth]{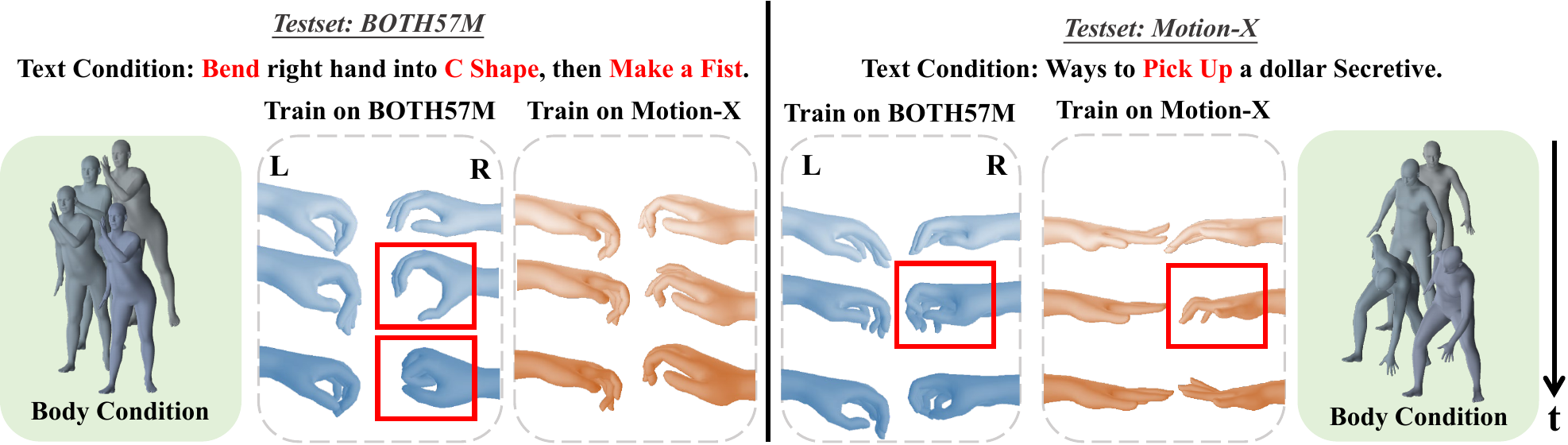}
    \vspace{-5mm}
    \caption{\textbf{Dataset Evaluation.} We train the BOTH2Hands algorithm on the training set of BOTH57M and Motion-X. Then sample on the test set of BOTH57M dataset (left) and Motion-X dataset (right). The poses that follow the conditions are circled \textcolor{red}{red}.
    }
    \label{fig:datasetevalu}
    \vspace{-10pt}
\end{figure*}

\subsection{Methods Evaluation} \label{sec: Methods Evaluation}
We compare our approach with several other methods in the task of generating hands based on textual and body conditions. We introduce latent text-to-motion methods T2M-GPT~\cite{zhang2023t2m} and MLD~\cite{chen2023executing}, diffusion-based method MDM~\cite{tevet2023human}, and body-conditioned motion synthesis method EgoEgo~\cite{li2023ego} for comparison. We align the input and output dimensions for unbiased comparison, keeping text and body conditions the same for all methods. In MLD, we employ two encoder-decoder structures for the body and hands. In the latent diffusion process, we merge the cleaned latent body token with the noisy hand token. The combined token is then denoised to predict latent hands, which are subsequently fed into the hand decoder. For T2M-GPT, we train an encoder-decoder structure to derive body features and then add up body and text tokens. For non-latent space methods, we directly concatenate the body conditions onto the noised hands as input and follow~\cite{tevet2023human} to add text conditions into it. All methods use the same implementation details as they presented. For the structures added to other methods, we keep dimensions the same as the original framework. For our method, all transformers consist of N=4 blocks, a latent dimension of 512, and 4 attention heads. We use a frozen CLIP-ViTL-14 model as the text encoder. As for other parameters, the diffusion timesteps are set to 1000 during training and inference; the AdamW optimizer is used with a fixed learning rate of $1e^{-4}$; and hyperparameter $w_{B}$ is set to 0.8, $w_{T}$ is set to 0.2; the motion blending process (method in Sec.\ref{sec: Cross-attention based gesture blending}) is performed 3 times. All the methods are trained with the BOTH57M training set on a single NVIDIA GeForce RTX 2080 Ti GPU for about 2 days. For more inference results, please refer to supplementary materials.
\begin{table}[h]
\centering
\renewcommand{\arraystretch}{1.2}
\resizebox{1\linewidth}{!}{
\begin{tabular}{l|l|c|c|c|c} 
\toprule 
Training Set           & Test Set & R precision$\uparrow$ & FID$\downarrow$ & Diversity$\rightarrow$ & MModality $\uparrow$ \\ 
\hline
\multirow{2}{*}{Real(GT)} & Motion-X &$0.044^{\pm.006}$&$0.353^{\pm.068}$&$3.224^{\pm.190}$& $-$   \\ %\cline{2-6}
                & BOTH57M  &$0.034^{\pm.020}$&$0.181^{\pm.012}$&  $3.980^{\pm.090}$  &  $-$\\ 
\hline
\multirow{2}{*}{Motion-X} & Motion-X & $\textbf{0.048}^{\pm.018}$&$\textbf{0.364}^{\pm.040}$ &  $3.203^{\pm.116}$  &  $\textbf{1.087}^{\pm.072}$\\ 
%\cline{2-6}
      & BOTH57M     &$0.026^{\pm.008}$ &$2.399^{\pm.076}$&$1.828^{\pm.200}$ &$0.581^{\pm.090}$            \\ 
\hline
\multirow{2}{*}{BOTH57M}     & Motion-X &$0.030^{\pm.002}$&$0.858^{\pm.024}$&$4.002^{\pm.054}$  & $1.259^{\pm.224}$  \\ 
                          & BOTH57M     & $\textbf{0.037}^{\pm.014}$&$\textbf{0.201}^{\pm.020}$&$3.969^{\pm.082}$  & $\textbf{1.312}^{\pm.034}$ \\
\hline
\end{tabular}
}
\vspace{-3mm}
\caption{\textbf{Cross-dataset comparisons} of BOTH57M and Motion-X. We train our pipeline on the training set of them, then evaluate the models on their test sets. Real(GT) means the GT data in the training set is used for evaluation. We use a 95\% confidence interval, approximated by the mean value plus or minus twice the standard deviation.}
\label{tab:datasetcompare}
\vspace{-15pt}
\end{table}

\begin{figure}[t]
    \vspace{-1.2ex}
    \centering
    \includegraphics[width=0.5\textwidth]{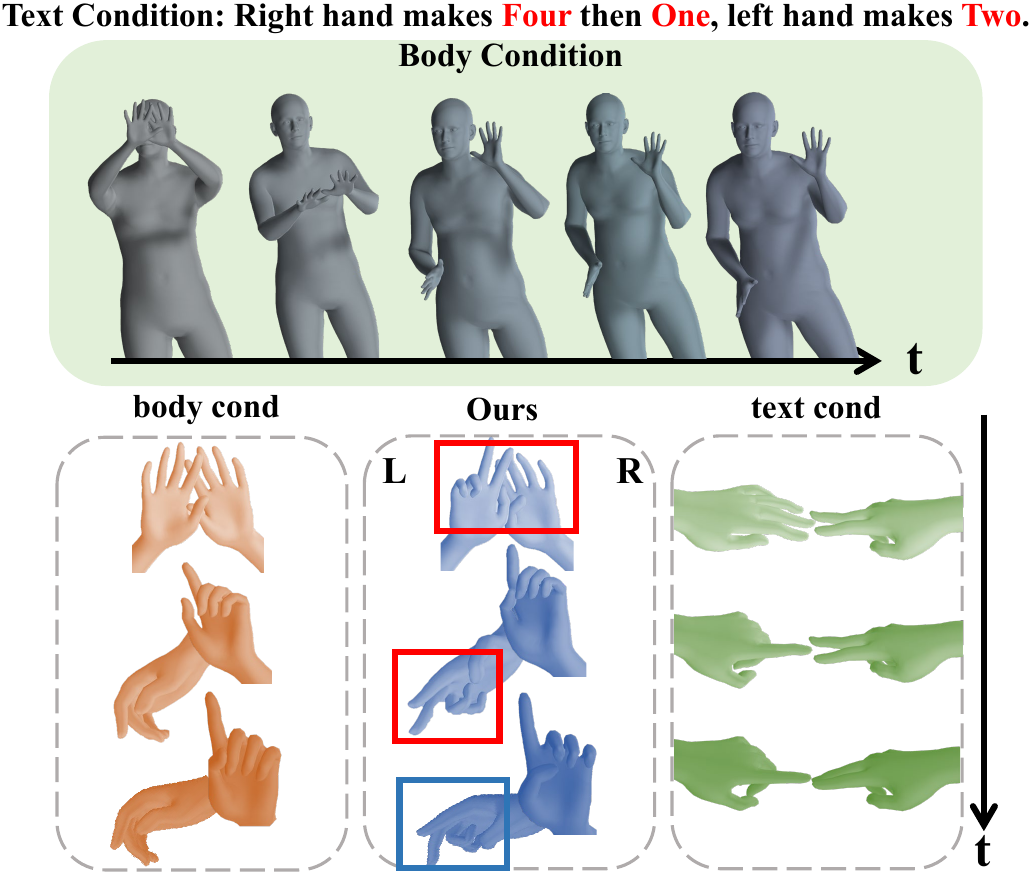}
    \vspace{-7.5mm}
    \caption{\textbf{Qualitative results of baseline comparison.} The motions following texts are circled \textcolor{red}{red}, the motions following body are circled \textcolor{blue}{blue}.
    } % follow
    \label{fig:baselinecomp}
    \vspace{-18pt}
\end{figure}

Following ~\cite{guo2022generating}, our evaluation metrics include Motion-retrieval precision (R Precision), Fréchet Inception Distance (FID)~\cite{heusel2017gans}, Multi-modal Distance (MM-Dist), Diversity and MultiModality (MModality). And we randomly split BOTH57M into the train (80\%), val (5\%), and test (15\%) set, adopting SMPLH as motion representation. Tab.~\ref{tab:methodcompare} presents detailed quantitative results from the same test set, showing our methods reconstructed motions closest to the real motion. Fig.~\ref{fig:methodeval} shows BOTH2Hands achieves good alignment between hand motions and conditions. Non-latent methods perform well on body conditions, but poorly on text conditions. Latent methods struggle with body conditions. Fig.~\ref{fig:baselinecomp} demonstrates our blending block beats our diffusion baseline. Text-conditioned hands lack body alignment, while body-conditioned hands fail to meet prompt requirements. Nevertheless, we also see marginal improvements in our evaluation results. This phenomenon implies that multi-conditioned generation performance may not be adequately reflected by widely used metrics such as R-precision, which are effective for single-conditioned generation evaluation. Two key reasons are listed below. First, in a full-body setting, hand motion constitutes a small portion, limiting the metric-based improvements. Second, the correlation between hand motions and text is non-linear, the metrics increase brought by enhancing hand motion is limited due to complicated hand-text alignment. We plan to explore more suitable metrics for future studies and hope our released code and dataset can serve as a solid foundation for such exploration.

\subsection{Dataset Evaluation} \label{sec: Dataset Evaluation}
To highlight the richness of the hand motions and the accuracy of text prompts, we compare our BOTH57M with Motion-X~\cite{lin2023motion}, the largest full-body motion dataset with text currently. We train BOTH2Hands method on training sets of Motion-X and BOTH57M separately. Then validate methods on respective test sets. The comparison results are presented in Tab.~\ref{tab:datasetcompare}. We add the GT data in the training set to the evaluation as the standard. The model trained on Motion-X training set performs well on the test set. However, the model trained on BOTH57M provides better alignment from text to hands, and its hand diversity is also better than the model trained on Motion-X. Fig.~\ref{fig:datasetevalu} shows our qualitative results. Given body and text conditions on test sets, our method trained on BOTH57M always performs better on text and body conditions. It also performs well on general motion prompts due to general motion annotations that contain hand descriptions in BOTH57M.  

\begin{figure}[t]
    \vspace{-1.6ex}
    \centering
    \includegraphics[width=0.47\textwidth]{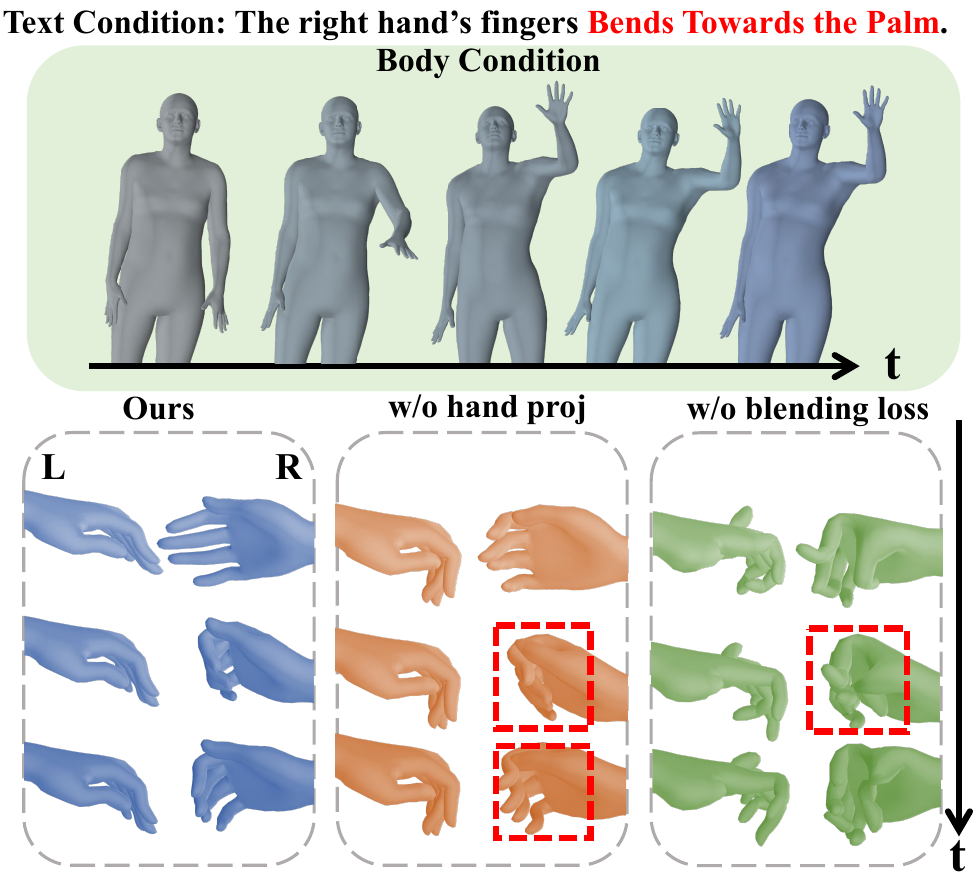}
    \vspace{-2.5mm}
    \caption{\textbf{Qualitative results of ablation study.} The errors of motions are circled \textcolor{red}{red}.
    }
    \label{fig:ablstudy}
    \vspace{-18pt}
\end{figure}

\begin{table}[h]
\vspace{-3mm}
\centering
\renewcommand\arraystretch{1.1}
\resizebox{1\linewidth}{!}{
\begin{tabular}{l|c|c|c} 
\hline
Method  & R Precision$\uparrow$ & FID$\downarrow$ & MM-Dist$\downarrow$  \\ 
\hline
GT    &  $0.034^{\pm.020}$           &  $0.181^{\pm.012}$      &   $1.391^{\pm.006}$            \\
Ours      & $\textbf{0.037}^{\pm.014}$            &  $\textbf{0.201}^{\pm.020}$     &    $\textbf{1.392}^{\pm.008}$           \\
w/o hand proj      &    $0.036^{\pm.014}$         &  $0.204^{\pm.026}$     &   $1.393^{\pm.008}$            \\
w/o blending loss      &    $0.034^{\pm.020}$         &  $0.210^{\pm.022}$     &  $1.392^{\pm.010}$             \\ 
\hline

\end{tabular}
}
\vspace{-3mm}
\caption{\textbf{Ablation study of BOTH2Hands algorithm.} Hand projection will fully improve the method results. We use a 95\% confidence interval, approximated by the mean value plus or minus twice the standard deviation.}
\label{tab:ablstudy}
\vspace{-15pt}
\end{table}

\subsection{Ablation Study}\label{sec: Ablation Study}
To validate the importance of hand projection and blending loss, we perform an ablation study by evaluating the effects of excluding these elements from the BOTH2Hands algorithm. Hand projection can be removed directly. But for blending loss, we choose linear distance loss as an alternative. Numerical results in Tab.~\ref{tab:ablstudy} indicate our method performs better with hand projection. This process allows the transformer to focus solely on the motion. Blending loss also highly improves hand motion quality, proving that learning the hand from previous output is effective. As mentioned in Sec.~\ref{sec: Dataset Evaluation}, marginal improvements found in the ablation study also suffer from the insensitivity of the existing metrics used. But obvious improvements shown in Fig.~\ref{fig:ablstudy} still prove the effectiveness of our hand projection and blending loss.

\section{Conclusion}
\label{sec:conclusion}
We introduce BOTH57M, the first comprehensive body-hand dataset that incorporates precise gestures and body movements, paired with meticulous finger-level hand annotations and body descriptions, which spans a variety of activities, consisting of 57.4 million frames in 8.31 hours, supplemented with 23,477 text annotations. Based on this dataset, we introduce BOTH2Hands, a robust algorithm designed to generate hand movements under two conditions: body movements and text prompts. Subsequently, we employ a cross-attention transformer for motion blending. We also conduct a series of detailed evaluations to demonstrate the robustness of our methods and the enhancement of our dataset for the two-hand generation task. 
We believe the BOTH57M could boost future exploration in multi-modal control and the analysis of human behavior.
\section{Acknowledgement}
\label{sec:acknowledgement}
This work was supported by National Key R\&D Program of China (2022YFF0902301), Shanghai Local college capacity building program (22010502800). We also acknowledge support from Shanghai Frontiers Science Center of Human-centered Artificial Intelligence (ShangHAI). 
{
    \small
    \bibliographystyle{ieeenat_fullname}
    \bibliography{arxiv}
}
\clearpage
\setcounter{page}{1}
\maketitlesupplementary

\renewcommand\thesection{\Alph{section}}
\setcounter{section}{0}
\section{More Details of BOTH57M}
\label{sec: More Details of BOTH57M}
In this section, we delve deeper into the specifics of our dataset. The following paragraphs introduce the camera dome setting, data reconstruction pipeline with data quality discussion, data text annotation, and data statics.
\subsection{Camera Dome Setting} 
\begin{figure*}[h]
    \centering
    \includegraphics[width=1.0\textwidth]{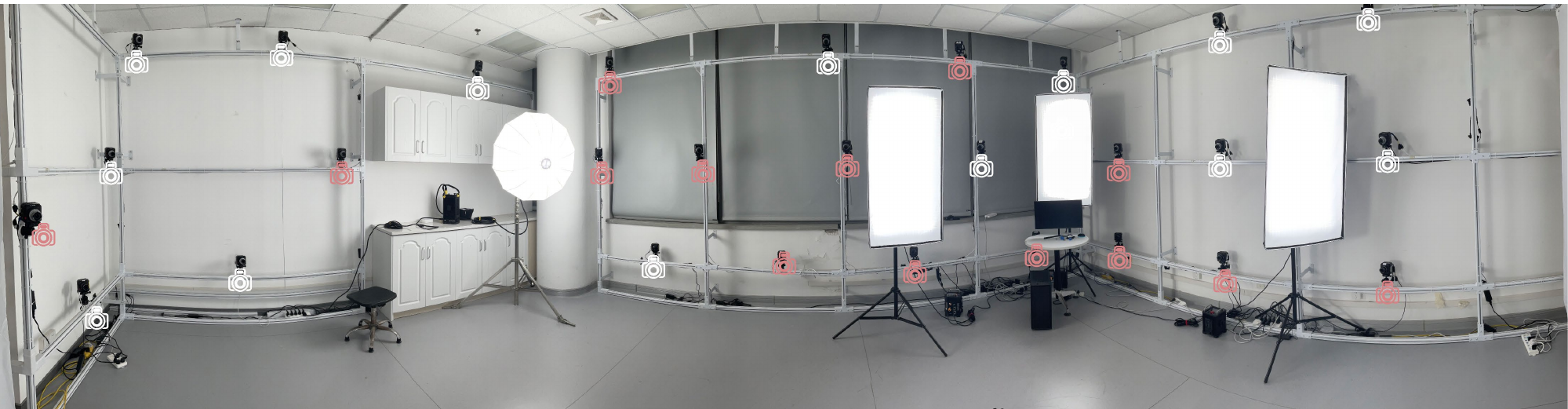}
    \vspace{-7mm}
    \caption{Dome setting, the red cameras are zoomed cameras for capturing hands, and other white cameras are wide-angle cameras for capturing the full body.}
    \label{fig:dome}
    \vspace{-3mm}
\end{figure*}
Our RGB cameras are mostly arranged evenly in three layers(low, medium, and high) at 270°. Other cameras are arranged on the highest layer at the remaining 90°. We zoom in on some of the cameras in order to ensure that they are sufficiently fine so that the motion of hands can be captured in detail. Moreover, we placed three 5500W fill lights in the dome to support hand motion capture with high-precision zoom that highly requires brightness. Fig.~\ref{fig:dome} shows the cameras and fill lights setting in the dome, the red cameras are zoomed in for capturing hands.

\subsection{Capture System Synchronization} 
In order to avoid motion blurring, the RGB cameras are set to work at 59.97 FPS with the resolution adjusted to 3840$\times$2160. Furthermore, cameras are available only if they are all turned on and set to the intended focal length.
When being captured, everyone is required to make a T-pose at the beginning and end of each shoot so that the initial orientation alignment of ground truth can be easily performed, i.e. the origin of the coordinate system is at the initial position of the person and the forward direction is towards the z-coordinate with up direction is towards the y-coordinate, using a right-handed system.

\subsection{Actor Requirements} 
During shooting, actors should perform motions in \textquotedblleft Dictionary of Gestures\textquotedblright ~\cite{caradec2018dictionary} excluding unfriendly gestures. They are required to expand the pose in the book into a complete body-hand motion sequence and perform it. Performers are encouraged to increase the diversity of hand gestures and body motions without changing the original meaning of the book.

\subsection{Data Reconstruction and Quality} 
\begin{figure}[h]
    \centering
    \includegraphics[width=0.45\textwidth]{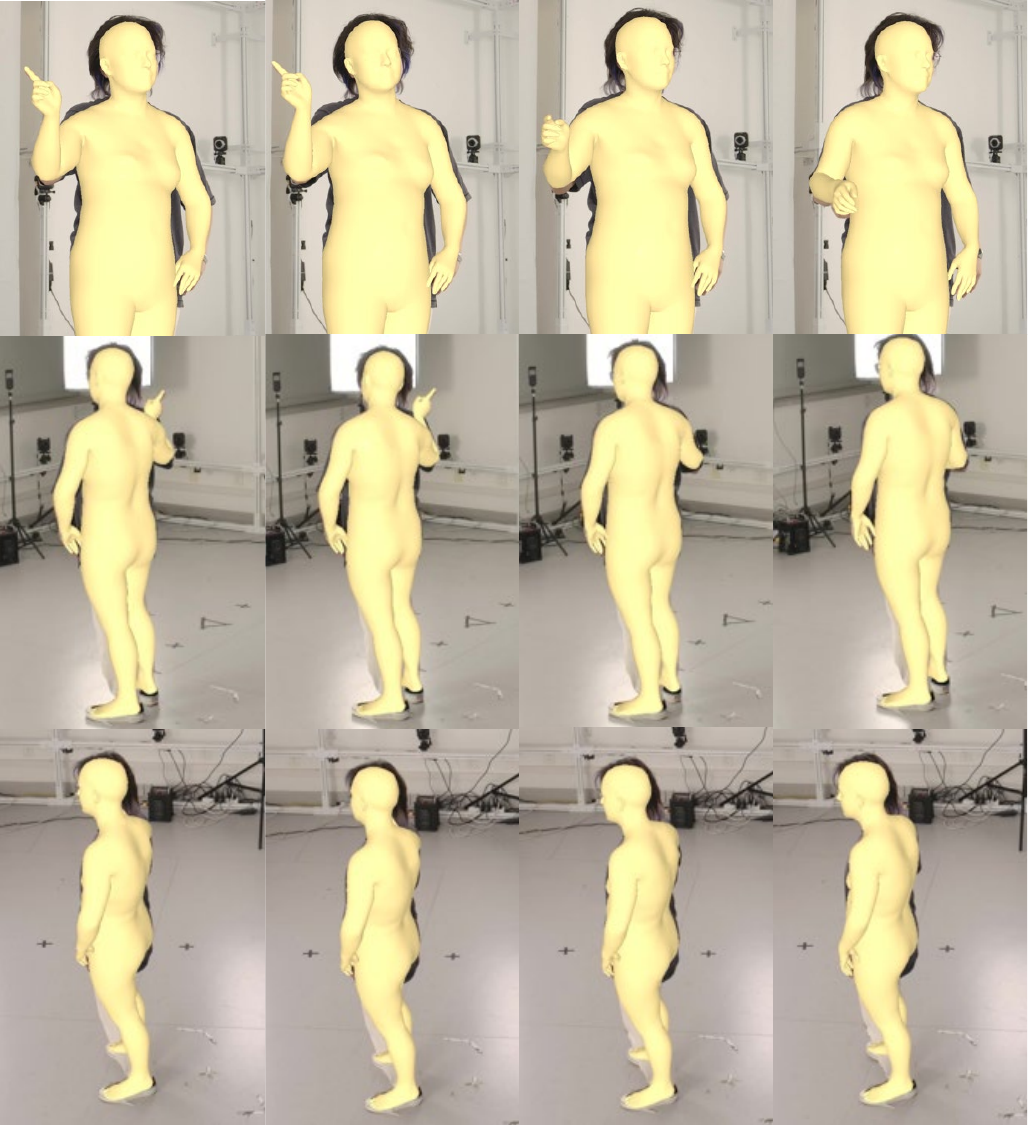}
    \caption{Overlays of BOTH57M in SMPLH. Three different views are presented, arranged from left to right in order of temporal variation.
    }
    \label{fig:overlay} 
\end{figure}

For data reconstruction, we adopt ViTPose~\cite{xu2022vitpose} for
predicting 2D body keypoints, and MediaPipe~\cite{lugaresi2019mediapipe} for estimating 2D hand keypoints from RGB images. Markerless
mocap then derives SMPLH parameters and 3D coordinates
from these 2D results. Fig.~\ref{fig:overlay} provides an overlay of our dataset. We are confident that our dataset exhibits sufficient full-body quality in 32-camera views with partial camera zoom-in for hand motion. We quantitatively validated our dataset using a subset with manually labeled 2D keypoints, triangulated for 3D joint coordinates as a gold standard. Our annotations show \textbf{31.4 mm} MPJPE and \textbf{25.2 mm} PA-MPJPE overall, with hand metrics at \textbf{6.51 mm} MPJPE and \textbf{3.97 mm} PA-MPJPE, which is comparable to the widely adopted datasets, e.g., 36.1mm for Human3.6M~\cite{ionescu2013human3} (full-body), 18.4mm for Grab~\cite{taheri2020grab} (body/hand marker-based), 2.78mm for InterHand~\cite{moon2020interhand2} (hand only), 33.5mm for Motion-X~\cite{lin2023motion} (pseudo-GT), etc. Nevertheless, we will release all the raw captured images, and welcome the community to further improve the tracking results.

\subsection{Data Text Annotation}
\begin{figure*}[h]
    \centering
    \includegraphics[width=0.9\textwidth]{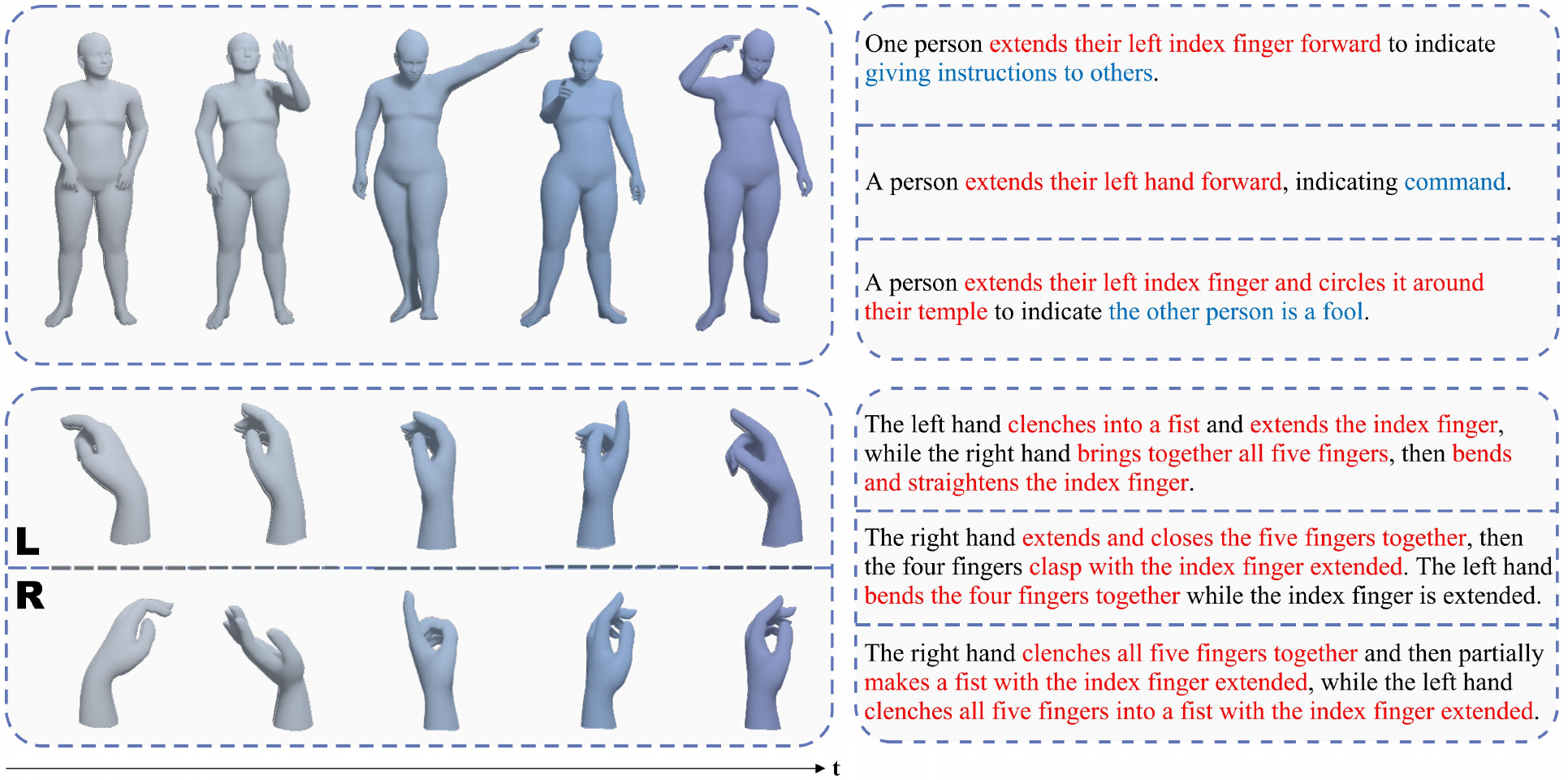}
    \caption{Detailed data structure of BOTH57M. We provide body annotations with general motion meaning, marked blue in the figure, and detailed finger-level annotations, the physical descriptions are marked red in the figure.}
    \label{fig:moredataset}
\end{figure*}

For data text annotation, we segment each motion into clips of 5 seconds. We also offer the data together with the book \textquotedblleft Dictionary of Gestures\textquotedblright ~\cite{caradec2018dictionary} to the annotators. For each motion sequence, three annotators are supposed to describe the motion according to the emotion from the corresponding poses in the book in \textquotedblleft Physical Description + Meaning of Action\textquotedblright \ format. \textquotedblleft Physical Description\textquotedblright \ means to describe the action precisely; \textquotedblleft Meaning of Action\textquotedblright \ means the emotion the action wants to express. For example, \textquotedblleft Put the five fingers of your right hand together in front of your right ear.\textquotedblright \ is a physical description, and \textquotedblleft Such an action is meant to express listening carefully.\textquotedblright \ is the meaning of the action. The other three annotators are required to focus on only hand movements and use detailed descriptions to explain how the fingers move. The annotations contain only physical descriptions such as \textquotedblleft Left-hand index finger straight, other fingers are naturally bent, right-hand thumb extended, other fingers in a fist\textquotedblright. Fig.~\ref{fig:moredataset} shows the motion of our dataset as well as the corresponding annotation in detail.

\begin{table}[t]
\centering
\renewcommand\arraystretch{1.1}
\resizebox{0.85\linewidth}{!}{
\begin{tabular}{lll}
\hline
Body part interact with & \# Frame & Time  \\ \hline
Head et al.             & 16.33M   & 2.33h \\
UpperBody et al.             & 5.64M   & 0.81h \\
Hand et al.             & 34.59M   & 5.01h \\
LowerBody et al.             & 1.27M   & 0.16h \\ \hline
Total                   & 57.83M   & 8.31h \\ \hline
\end{tabular}
}
\vspace{-3mm}
\caption{BOTH57M Subjects.}
\label{tab:DataStatics}
\vspace{-7mm}
\end{table}

\subsection{Data Statics}
BOTH57M is the only dataset that provides hybrid and detailed annotations of both body and hands at present. It consists of 1,384 motion clips and 57.4M frames, with 23,477 manually annotated motions and a rich vocabulary of 4,140 words. The data-capturing system includes 32 synchronized high-resolution RGB cameras with 59.97 FPS and 3840×2160 resolution, which are capable of full body capture tasks. As for subjects of the dataset, we follow \textquotedblleft Dictionary of Gestures\textquotedblright \ to scientifically split the dataset into 36 subjects according to the body part that hands interact with, a statistic is shown in Tab.\ref{tab:DataStatics}. 

\begin{table*}
\centering
\caption{Quantitative evaluation of our condition hyperparameter setting, red and blue results indicate the best result and the second best result. We use a 95\% confidence interval, approximated by the mean value plus or minus twice the standard deviation.}
\begin{tabular}{cccccccc} 
\hline
 {$w_B/w_T$} & \multicolumn{3}{c}{R Precision$\uparrow$} & FID$\downarrow$ & MM-Dist$\downarrow$ & Diversity$\rightarrow$ & MModality$\uparrow$     \\ 
\cline{2-4}
          & Top1        & Top2        & Top3        &             &             &             &              \\ 
\hline
$1.0/{0.0}$ & $0.036^{\pm0.026}$ & $0.064^{\pm0.034}$ & $0.094^{\pm0.038}$ & $0.201^{\pm0.018}$ & $1.402^{\pm0.016}$ & $3.983^{\pm0.062}$ & $1.262^{\pm0.138 }$ \\
$0.9/{0.1}$ & $0.032^{\pm0.022}$ & $0.063^{\pm0.028}$ & $0.093^{\pm0.034}$ & $\textcolor{red}{\textbf{0.198}}^{\pm0.022}$ & $1.402^{\pm0.012}$ & $3.968^{\pm0.074}$ & $1.298^{\pm0.124 }$ \\
$0.8/{0.2}$ & $0.037^{\pm0.014}$ & $\textcolor{blue}{\textbf{0.075}}^{\pm0.020}$ & $\textcolor{red}{\textbf{0.115}}^{\pm0.028}$ & $0.201^{\pm0.020}$ & $\textcolor{blue}{\textbf{1.392}}^{\pm0.008}$ & $3.969^{\pm0.082}$ & $\textcolor{red}{\textbf{1.312}}^{\pm0.034 }$ \\
$0.7/{0.3}$ & $0.031^{\pm0.020}$ & $0.060^{\pm0.024}$ & $0.088^{\pm0.015}$ & $\textcolor{blue}{\textbf{0.199}}^{\pm0.020}$ & $1.404^{\pm0.008}$ & $3.968^{\pm0.060}$ & $\textcolor{blue}{\textbf{1.303}}^{\pm0.104 }$ \\
$0.6/{0.4}$ & $0.030^{\pm0.024}$ & $0.059^{\pm0.028}$ & $0.089^{\pm0.032}$ & $0.210^{\pm0.018}$ & $1.405^{\pm0.012}$ & $3.980^{\pm0.060}$ & $1.291^{\pm0.126 }$ \\
$0.5/{0.5}$ & $0.032^{\pm0.012}$ & $0.069^{\pm0.024}$ & $0.096^{\pm0.032}$ & $0.223^{\pm0.026}$ & $1.401^{\pm0.012}$ & $3.939^{\pm0.048}$ & $1.267^{\pm0.246 }$ \\
$0.4/{0.6}$ & $\textcolor{blue}{\textbf{0.038}}^{\pm0.010}$ & $0.074^{\pm0.016}$ & $0.104^{\pm0.024}$ & $0.237^{\pm0.038}$ & $\textcolor{red}{\textbf{1.391}}^{\pm0.010}$ & $3.939^{\pm0.070}$ & $1.250^{\pm0.136 }$ \\
$0.3/{0.7}$ & $0.038^{\pm0.016}$ & $0.074^{\pm0.026}$ & $0.110^{\pm0.026}$ & $0.231^{\pm0.028}$ & $1.392^{\pm0.014}$ & $3.927^{\pm0.056}$ & $1.287^{\pm0.112 }$ \\
$0.2/{0.8}$ & $0.035^{\pm0.012}$ & $0.069^{\pm0.014}$ & $0.108^{\pm0.016}$ & $0.235^{\pm0.026}$ & $1.392^{\pm0.010}$ & $3.922^{\pm0.066}$ & $1.231^{\pm0.130 }$ \\
$0.1/{0.9}$ & $\textcolor{red}{\textbf{0.041}}^{\pm0.020}$ & $\textcolor{red}{\textbf{0.079}}^{\pm0.020}$ & $\textcolor{blue}{\textbf{0.113}}^{\pm0.032}$ & $0.233^{\pm0.026}$ & $1.392^{\pm0.010}$ & $3.934^{\pm0.052}$ & $1.223^{\pm0.268 }$ \\
$0.0/{1.0}$ & $0.033^{\pm0.010}$ & $0.067^{\pm0.022}$ & $0.104^{\pm0.030}$ & $0.237^{\pm0.028}$ & $1.393^{\pm0.008}$ & $3.972^{\pm0.058}$ & $1.266^{\pm0.130 }$ \\
\hline
\end{tabular}
\label{tab:paracompare}
\vspace{-15pt}
\end{table*}

\section{More Experiment Results}
In this section, we provide more experiment visualization results and an extra hyperparameter experiment finding suitable $w_B$ and $w_T$ settings.
\subsection{Hyperparameter experiments} 
We establish 11 sets of hyperparameter experiments with an interval of 0.1, starting at 0, under the premise that $w_B + w_T = 1$. We limit $w_B + w_T = 1$ to ensure the cross-attention transformer layer returns the same order of magnitude of loss as the previous diffusion process. Tab.~\ref{tab:paracompare} shows detailed results under different hyperparameter settings. The evaluation metrics are composed of the following sections: Motion-retrieval precision (R Precision), which we adhere to the top 3 results, is used to measure the alignment degree between conditions and motions. Fréchet Inception Distance (FID)~\cite{heusel2017gans} is employed to evaluate the feature distribution between the generated actions and the ground truth motion.  Multi-modal Distance (MM-Dist) calculates the distance between hand motions and body text conditions. Diversity assesses the richness of generation motion by calculating the variance of data. Multimodality (MModality) measures the variety of generated motions under the same input conditions. The experimental results demonstrate that utilizing solely the body or text as the condition, corresponding to setting $w_T=0$ or $w_B=0$ in the experiment, can not get the best results. As the weight of the text condition $w_T$ increases, the alignment performance (R Precision) between motion and condition improves, but the authenticity of the motion (FID) deteriorates. As the weight of the body condition $w_B$ increases, the authenticity of the motion (FID) improves, but the Multi-modal Distance (MM-Dist) also increases, indicating that the distance between the generated hand motion result and the condition has grown farther in latent space. Under the experiment results, we decide to use a group with a good match degree between motion and condition, and relatively realistic motion, specifically setting the hyperparameters of the pipeline as $w_B=0.8$ and $w_T=0.2$.

\subsection{More evaluation results} 
\begin{figure*}[h]
    \centering
    \includegraphics[width=0.85\textwidth]{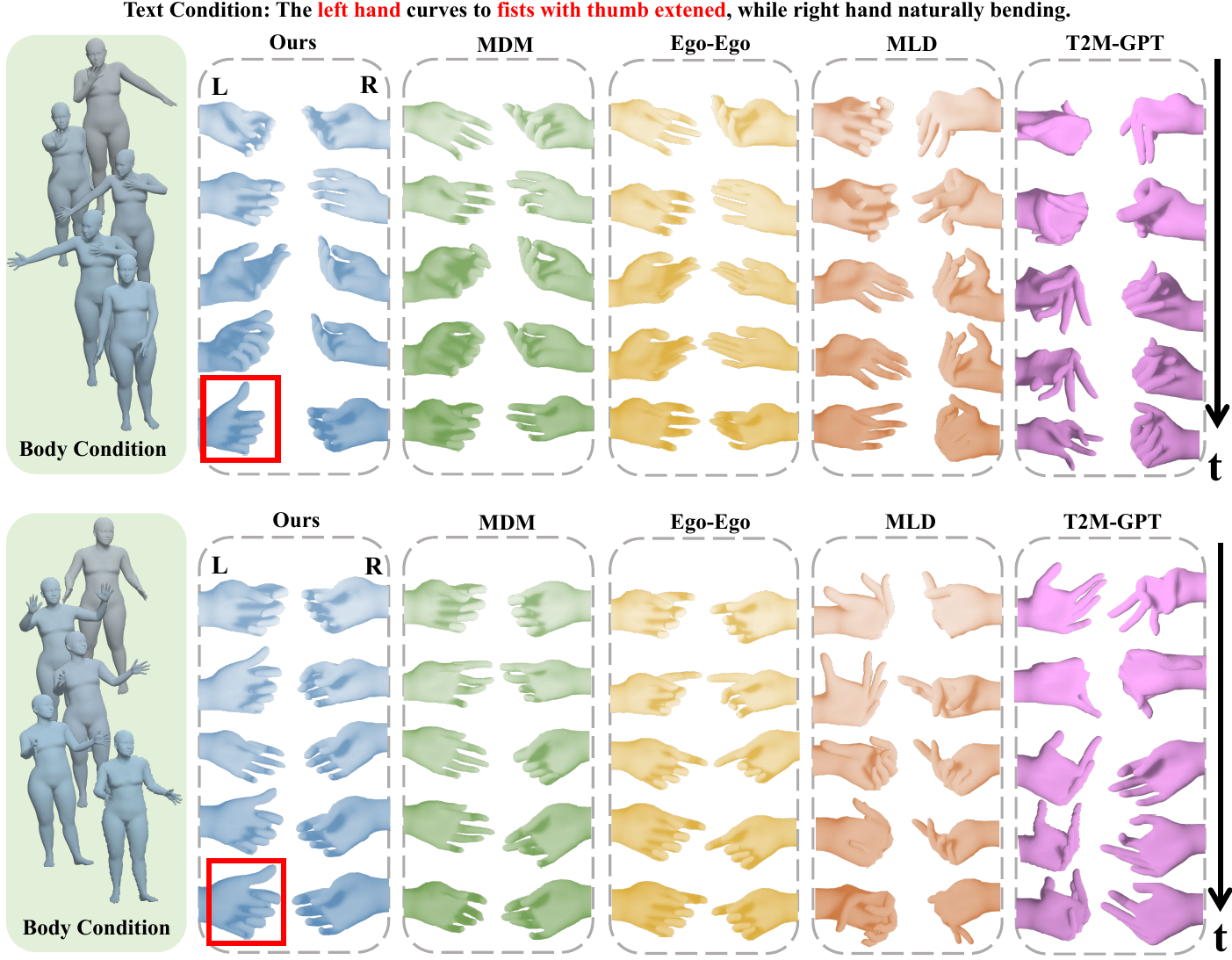}
    \caption{More method evaluation results. Our method performs well under different body conditions.}
    \label{fig:moremethodeval}
\end{figure*}

\begin{figure*}[h]
    \centering
    \includegraphics[width=0.85\textwidth]{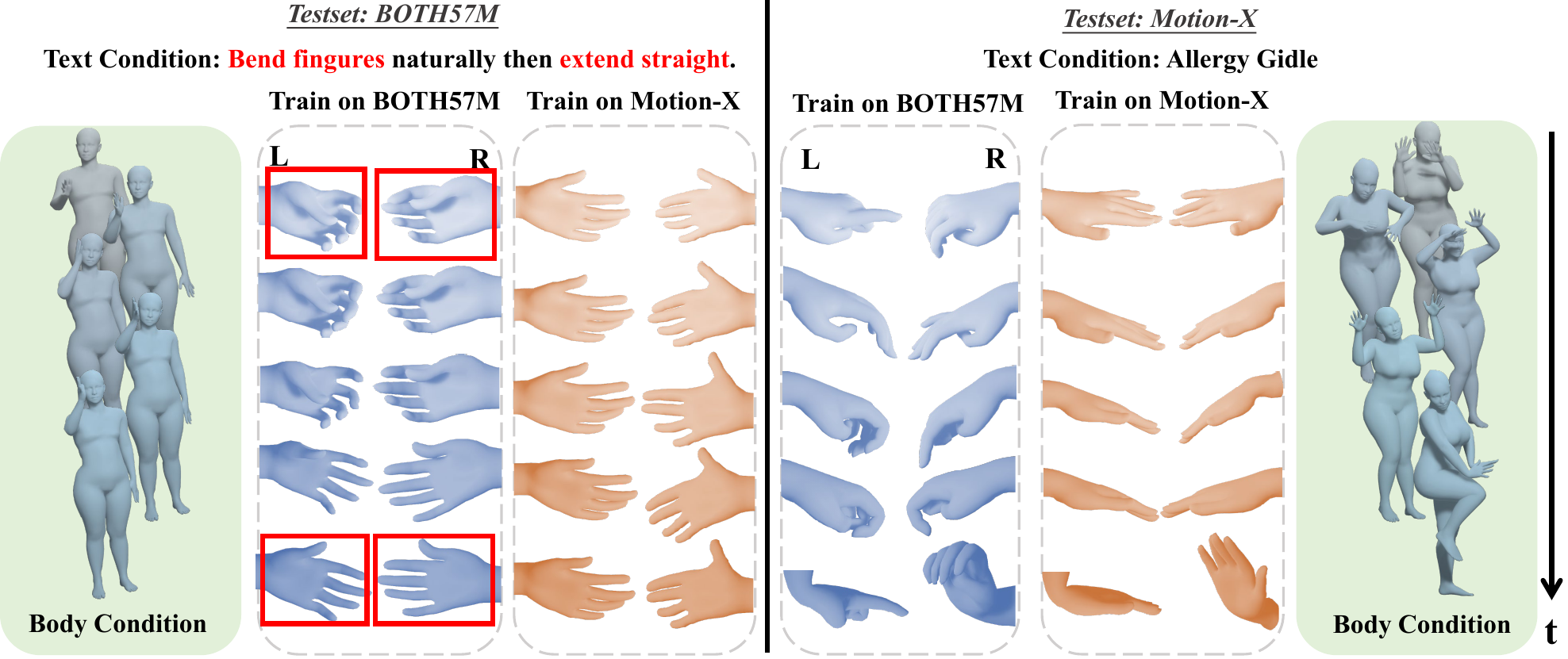}
    \caption{More dataset evaluation results. BOTH57M continues to exhibit robust generalization even under challenging motion and text conditions.}
    \label{fig:moredataseteval}
\end{figure*}

\begin{figure*}[h]
    \centering
    \includegraphics[width=\textwidth]{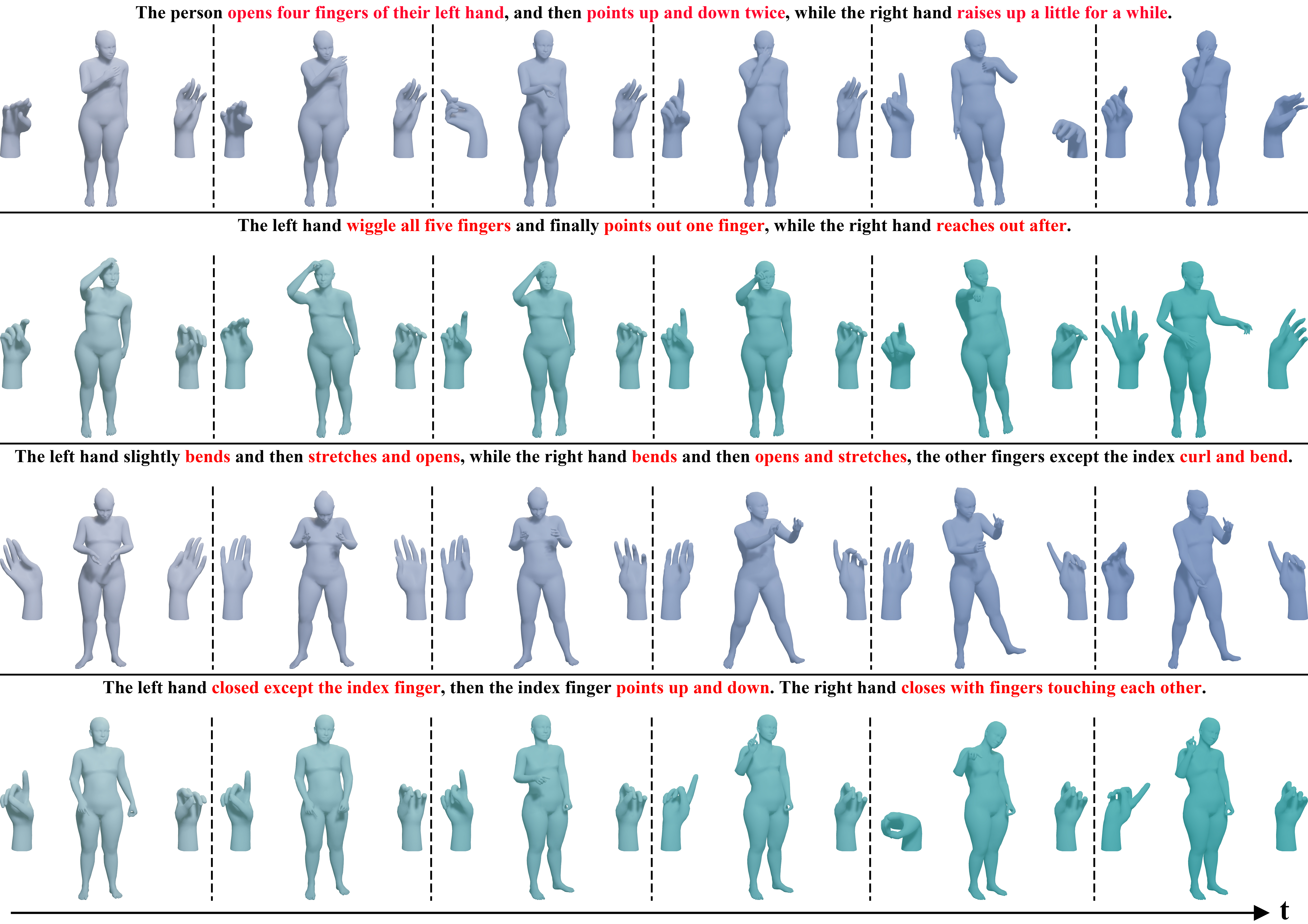}
    \vspace{-3mm}
    \caption{Qualitative results generated by our BOTH2Hands algorithm. We showcase text prompts at the top of each motion. From left to right, the temporal order is indicated.}
    \label{fig:gallery}
\end{figure*}

We present more visualization results of comparing methods to further validate the effectiveness of the BOTH2Hands algorithm. As shown in Fig.~\ref{fig:moremethodeval}, under the same text condition but different body conditions, our method can still generate hand results that correspond to the text and body. We also illustrate more cross-validation visualization results. As depicted in Fig.~\ref{fig:moredataseteval}, under the challenge of more vague text descriptions and professional dance movements, our dataset can still provide lively hand gestures that align with text controls, demonstrating the broad generalization capabilities of the BOTH57M dataset. Finally, we present a gallery of our inference results for the text/body-to-hand task in Fig.~\ref{fig:gallery}.

\begin{figure*}[t!]
\vspace{-5mm}
    \centering
    \includegraphics[width=\textwidth]{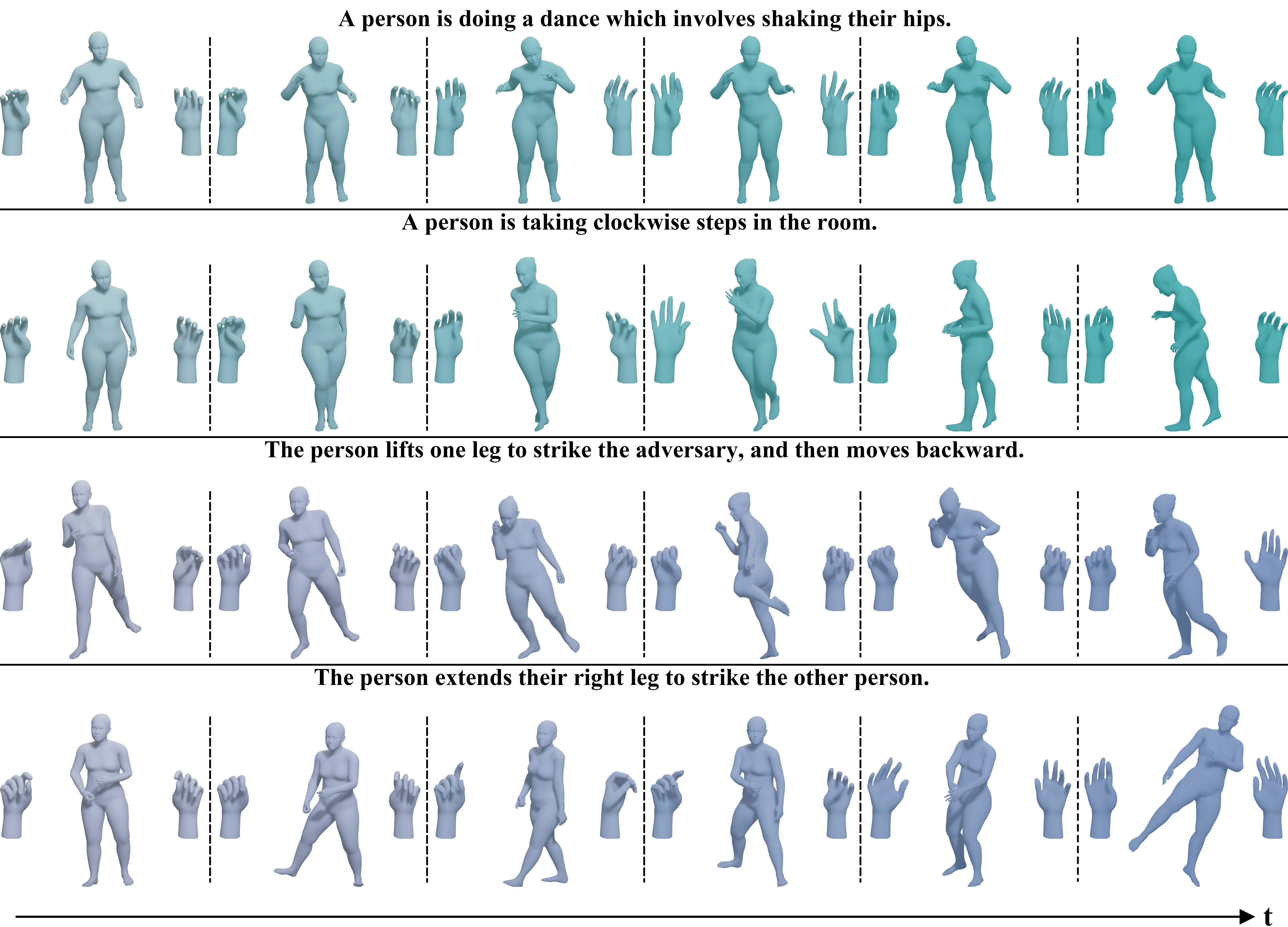}
    \vspace{-3mm}
    \caption{Inference on HumanML3D~\cite{guo2022generating} and InterHuman~\cite{liang2023intergen}. We pretrain our BOTH2Hands algorithm on BOTH57M, and inference two-hand motions using text and body conditions of HumanML3D and InterHuman. Our results faithfully match the daily motion in HumanML3D and challenging professional motions in InterHuman.}
    \label{fig:application}
\end{figure*}
\section{Applications}
Our BOTH2Hands algorithm, trained on BOTH57M dataset, has demonstrated excellent generalization capabilities even on other human body datasets that only offer body-level descriptions. We conducted experiments by inputting body and text data from the HumanML3D~\cite{guo2022generating} dataset into BOTH2Hands, successfully generating hand movements that are remarkably vivid. In a bid to challenge the generation of hand movements in more professional actions, we proceed to perform inference on InterHuman~\cite{liang2023intergen}, a dataset encompassing professional movements such as ballet, boxing, and more. Fig.~\ref{fig:application} illustrates how our method can effectively augment data for existing body and text datasets.

\section{Limitation and Broader Impact}
In this section, we discuss the border impact and limitations.

\begin{figure*}[t]
    \centering
    \includegraphics[width=1.0\textwidth]{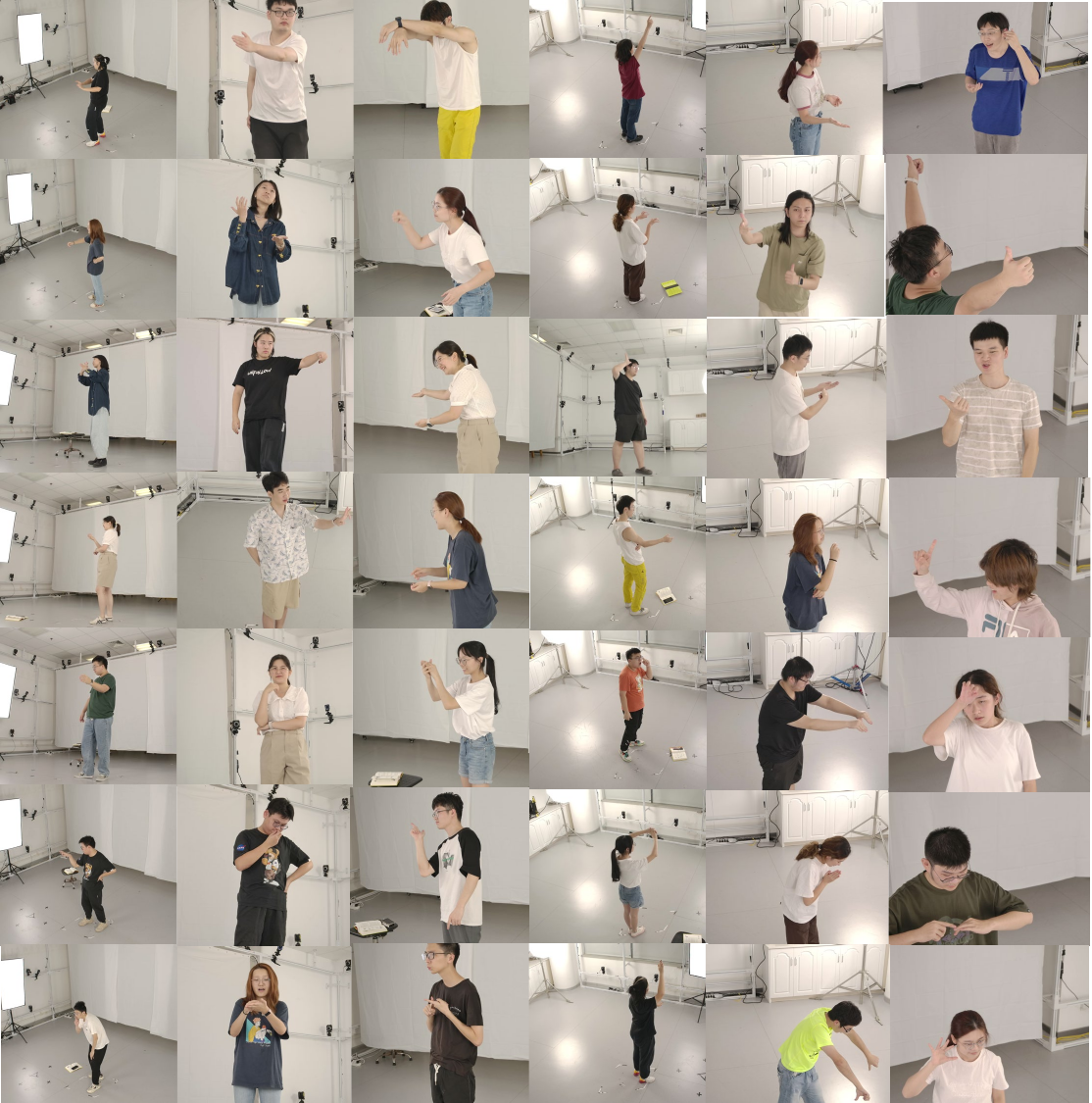}
    \vspace{-7mm}
    \caption{Data examples captured by our dome with 32 synchronized high-resolution RGB cameras. Our dataset includes a variety of body-hand motions under various daily scenes.}
    \label{fig:realman}
    \vspace{-5mm}
\end{figure*}

\myparagraph{Broader Impact.} 
We are the first to propose generating fine-grained two-hand motions through implicit control from real-world body input and user-defined explicit text control. This provides a direction that can be studied for user-customized motion generation. Furthermore, the pre-trained BOTH2Hands, based on the hand synthesis through the parallel diffusion structure and large-scale motion data training on the BOTH57M, can enhance data for large body motion datasets with textual descriptions like HumanML3D~\cite{guo2022generating} and InterHuman~\cite{liang2023intergen}. The research also contributes to balancing control signals during the multi-signal control diffusion process. Furthermore, text-controllable and body-aligned hand synthesis has potential in many practical application scenarios, such as Virtual Reality (VR), Augmented Reality (AR) games, animations, and human communication studies, etc. Fig.~\ref{fig:realman} provides the gallery of data examples captured by our dome with 32 synchronized high-resolution RGB cameras.

\myparagraph{Limitation.} 
While this work has achieved substantial advancements in the novel task of generating hand motions in alignment with body and text and offering an intuitive control method, it has limitations. First, the fine-grained spatial control provided by text descriptions is sufficient but lacks temporal alignment methods. Second, we focused on the connection between hand gestures and the body as a whole but did not take into account how different parts of the body might separately affect the hands. Third, more suitable metrics that can precisely reflect the alignment between hand, body, text, and other complex conditions should be presented, we welcome the community to focus on it and hope our contribution will push the field forward. Finally, although BOTH2Hands can generate vivid hand poses, it remains challenging to determine whether a two-handed interaction gesture is generated when the wrists are close enough. Hand-interaction poses represent a hugely different meaning from poses that do not interact. Future work may focus on aligning text control temporally and explore how individual body parts relate to hand movements, especially hand-to-hand interaction. Improved metrics are needed to better represent hand and multi-condition alignment.

\end{document}